\documentclass[lettersize,journal]{IEEEtran}
\usepackage[caption=false,font=normalsize,labelfont=sf,textfont=sf]{subfig}
\usepackage{stfloats}
\usepackage{verbatim}
\usepackage{cite}
\usepackage{amsmath,amssymb,amsfonts}
\usepackage{graphicx}
\usepackage{algorithm}
\usepackage{algpseudocode}
\usepackage{threeparttable}
\usepackage{graphics}
\usepackage{epsfig}
\usepackage{textcomp}
\usepackage{xcolor}
\usepackage{amssymb}
\usepackage{booktabs}
\usepackage{multirow}
\usepackage{makecell}
\usepackage{bbding}  
\usepackage{pifont}
\usepackage{multicol}
\usepackage{hyperref}
\usepackage{colortbl}
\definecolor{dark-green}{HTML}{228B22}
\definecolor{dark-blue}{HTML}{1976D2}
\definecolor{dark-purple}{HTML}{8d4bbb}
\definecolor{dark-red}{HTML}{D63C3C}
\definecolor{pink}{HTML}{fae6e9}
\definecolor{n1}{HTML}{ff9999}
\definecolor{n2}{HTML}{FFCC99}
\definecolor{n3}{HTML}{FFFF99}
\newcommand{\cmark}{\ding{51}}
\newcommand{\xmark}{\ding{55}}

\hypersetup{
    colorlinks=true,
    linkcolor=dark-red,
    filecolor=magenta,      
    urlcolor=dark-red,
    citecolor=dark-blue,
}
\usepackage{url}
\usepackage{mathtools}

\newcommand{\shh}[1]{\textcolor{black}{#1}}

\begin{document}

\title{Trio-ViT: Post-Training Quantization and Acceleration for Softmax-Free Efficient Vision Transformer}

\author{Huihong Shi, Haikuo Shao, Wendong Mao, and Zhongfeng Wang,~\IEEEmembership{Fellow,~IEEE}

\thanks{This work was supported in part by the National Key R\&D Program of China under Grant 2022YFB4400600 and in part by the Shenzhen Science and Technology Program 2023A007.}
\thanks{Huihong Shi and Haikuo Shao are with the School of Electronic Science and Engineering, Nanjing University, Nanjing, China (e-mail: \{{shihh, hkshao}\}@smail.nju.edu.cn).}
\thanks{Wendong Mao is with the School of Integrated Circuits, Sun Yat-sen University, Shenzhen, China (e-mail:maowd@mail.sysu.edu.cn).}
\thanks{Zhongfeng Wang is with the School of Electronic Science and Engineering, Nanjing University, and the School of Integrated Circuits, Sun Yat-sen University (email: zfwang@nju.edu.cn).}
\thanks{Correspondence should be addressed to Wendong Mao and Zhongfeng Wang.}}

\maketitle
\begin{abstract}
Motivated by the huge success of Transformers in the field of natural language processing (NLP), Vision Transformers (ViTs) have been rapidly developed and achieved remarkable performance in various computer vision tasks. However, their huge model sizes and intensive computations hinder ViTs' deployment on embedded devices, calling for effective model compression methods, such as quantization. 
Unfortunately, due to the existence of hardware-unfriendly and quantization-sensitive non-linear operations, particularly {Softmax}, it is non-trivial to completely quantize all operations in ViTs, yielding either significant accuracy drops or non-negligible hardware costs. 
In response to challenges associated with \textit{standard ViTs}, we focus our attention towards the quantization and acceleration for \textit{efficient ViTs}, which not only eliminate the troublesome Softmax but also integrate linear attention with low computational complexity, and propose \emph{Trio-ViT} accordingly. 
Specifically, at the algorithm level, we develop a {tailored post-training quantization engine} taking the unique activation distributions of Softmax-free efficient ViTs into full consideration, aiming to boost quantization accuracy. 
Furthermore, at the hardware level, we build an accelerator dedicated to the specific Convolution-Transformer hybrid architecture of efficient ViTs, thereby enhancing hardware efficiency.
Extensive experimental results consistently prove the effectiveness of our Trio-ViT framework. {Particularly, we can gain up to $\uparrow$$\mathbf{3.6}\times$, $\uparrow$$\mathbf{5.0}\times$, and $\uparrow$$\mathbf{7.3}\times$ FPS under comparable accuracy over state-of-the-art ViT accelerators, as well as $\uparrow$$\mathbf{6.0}\times$, $\uparrow$$\mathbf{1.5}\times$, and $\uparrow$$\mathbf{2.1}\times$ DSP efficiency.}
Codes are available at \url{https://github.com/shihuihong214/Trio-ViT}.
\end{abstract}


\section{{{Introduction}}}
\label{sec:intro}
\IEEEPARstart{T}{hanks} to the powerful global information extraction capability of self-attention mechanism, Transformers have achieved great success in various natural language processing (NLP) tasks \cite{Vaswani2017AttentionIA, Devlin2019BERTPO, OpenAI2023GPT4TR}.
This success has inspired the rapid development of Vision Transformers (ViTs) \cite{vit, deit}, which have gained increasing attention in the field of computer vision and shown superior performance compared to their convolution-based counterparts.
However, their enormous model sizes and intensive computations challenge the deployment of ViTs on embedded/mobile devices, where both memory and computing resources are limited.
For example, ViT-Large \cite{vit} contains $307$M parameters and yields $190.7$G FLOPs during inference.
Thus, effective model compression techniques are highly desired to facilitate ViTs' real-world applications. 

Among them, model quantization \cite{Li2021BRECQPT, Jacob2017QuantizationAT, Yao2020HAWQV3DN, p2-vit} stands out as one of the most effective and widely adopted compression methods. It converts floating-point weights/activations into integers, leading to a reduction in both memory consumption and computational costs during inference.
Unfortunately, due to the existence of non-linear operations, including LayerNorm (LN), GELU, and especially Softmax, which are not only hardware unfriendly but also quantization-sensitive, ViTs are difficult to be fully quantized, yielding either significant accuracy drops or notable hardware overhead \cite{Lin2021FQViTPQ, Yuan2021PTQ4ViTPQ}. 
To solve these challenges, several efforts \cite{Lin2021FQViTPQ, Li2022IViTIQ, Yuan2021PTQ4ViTPQ} have been devoted. 
For example, FQ-ViT \cite{Lin2021FQViTPQ} identifies extreme inter-channel variations in LN's inputs and excessive non-uniform distributions in attention maps, and proposes Power-of-Two Factor (PTF) and Log-Int-Softmax (LIS) for LN and Softmax quantization, respectively. 
Additionally, I-ViT \cite{Li2022IViTIQ} develops innovative lightweight dyadic arithmetic methods to approximate ViTs' non-linear operations, thus achieving integer-only inference.
Despite their effectiveness, they are dedicated to the quantization for standard ViTs while overlooking the inherent quantization and acceleration opportunities within efficient ViTs \cite{cai2022efficientvit, Han2023FLattenTV, You2022CastlingViTCS}, where the vanilla Softmax-based attention with quadratic computational complexity is typically traded with more efficient \emph{Softmax-free attentions} that exhibit \emph{linear} computational complexity. 
To close this gap, we redirect our focus towards the exploration of effective quantization and acceleration for efficient ViTs, aiming to fully unleash their potential algorithmic benefits to win both accuracy and hardware efficiency, i.e., {\textbf{(i)}} the Softmax-free property to boost the achievable quantization performance and {\textbf{(ii)}} the linear complexity characteristic of attentions to enhance inference efficiency.

In addition to the algorithm level, various works have built dedicated accelerators to boost ViTs' hardware efficiency from the hardware perspective \cite{Sun2022VAQFFA, Li2022AutoViTAccAF, You2022ViTCoDVT}. For instance, Auto-ViT-Acc \cite{Li2022AutoViTAccAF} adopts mixed quantization schemes, i.e., fixed-point and power-of-two, to quantize ViTs, and develops a dedicated accelerator to fully leverage the computational resources available on FPGAs.
Moreover, ViTCoD \cite{You2022ViTCoDVT} proposes pruning and polarization techniques to transform ViTs' attention maps into denser and sparser variants, and then develops a dedicated accelerator incorporating both dense and sparse engines to simultaneously execute the above two workloads.
Despite the superiority of the above accelerators in enhancing hardware efficiency, they are dedicated to standard ViTs and fall short in fully accelerating efficient ViTs \cite{cai2022efficientvit, Han2023FLattenTV, You2022CastlingViTCS}, which are typically characterized by {(i)} Softmax-free linear attentions and {(ii)} Convolution-Transformer hybrid architectures.
Specifically, it has been widely verified that the computational complexity reduction of linear attentions will yield a degradation in their local feature extraction ability, thus necessitating extra compensation components such as convolutions \cite{cai2022efficientvit, Han2023FLattenTV}. This results in hybrid architectures for efficient ViTs that comprise both convolutions and Transformer blocks, thus calling for dedicated accelerators to unleash their potential benefits. 

To grasp the inherent quantization and acceleration opportunities in efficient ViTs, we make the following contributions: 
\begin{itemize}
    \item We propose \textbf{\textit{Trio-ViT}}, a post-training quantization and acceleration framework for efficient Vision Transformers (ViTs) via algorithm and hardware co-design. To the best of our knowledge, this is the first work dedicated to the quantization and acceleration of efficient ViTs.
    
    \item At the algorithm level, we conduct a comprehensive
    analysis of distinct activations of Softmax-free efficient ViTs and unveil specific quantization challenges. Then, we develop a tailored post-training quantization engine that incorporates several novel strategies, including \textbf{\textit{channel-wise migration, filter-wise shifting}}, and \textbf{\textit{log2 quantization}}, to address the involved challenges with boosted quantization accuracy. 
    
    \item At the hardware level, we advocate a \textbf{\textit{hybrid design}} incorporating multiple computing cores to effectively support various operation types in the Convolution-Transformer hybrid architecture of efficient ViTs.
    Besides, we propose a \textbf{\textit{pipeline architecture}} to facilitate both inter- and intra-layer fusions, thus enhancing hardware utilization and easing the bandwidth requirement.
   
    \item Extensive experiments and ablation studies consistently validate the effectiveness of our Trio-ViT framework. For example, we can offer up to $\uparrow$$\mathbf{3.6}\times$, $\uparrow$$\mathbf{5.0}\times$, and $\uparrow$$\mathbf{7.3}\times$ FPS with comparable accuracy over state-of-the-art (SOTA) ViT acceleration frameworks. Besides, we can achieve up to $\uparrow$$\mathbf{6.0}\times$, $\uparrow$$\mathbf{1.5}\times$, and $\uparrow$$\mathbf{2.1}\times$ DSP efficiency. It is expected that our work can open up an exciting perspective for the quantization and acceleration of Softmax-free efficient ViTs.
\end{itemize}

\textcolor{black}{The rest of this paper is organized as follows}: we first introduce related works in Sec. \ref{sec:related_work} and preliminaries in Sec. \ref{sec:Preliminaries}; Then, we illustrate Trio-ViT's post-training quantization engine and dedicated accelerator in Sec. \ref{sec:quantization_alg} and Sec. \ref{sec:Accelerator}, respectively; Furthermore, extensive experiments and ablation studies consistently demonstrate Trio-ViT's effectiveness in Sec. \ref{sec:experiments};
Finally, Sec. \ref{sec:conclusion} summarizes this paper.

\section{Related Works}
\label{sec:related_work}
\subsection{Model Quantization for Vision Transformers (ViTs)}
Model quantization, which represents floating-point weights and activations with integers without modifying model architectures, is a generic compression solution.
It can be roughly categorized into two approaches: quantization-aware training (QAT) and post-training quantization (PTQ).
Specifically, QAT \cite{Dong2019HAWQHA, Yao2020HAWQV3DN, Shen2019QBERTHB, Li2022IViTIQ} involves weight fine-tuning to facilitate quantization, yielding higher accuracy or lower quantization bit.
In contrast, PTQ \cite{Lin2021FQViTPQ, Yuan2021PTQ4ViTPQ, Liu2021PostTrainingQF, Xiao2022SmoothQuantAA, p2-vit}, which eliminates resource-intensive fine-tuning and streamlines models' deployment, has recently gained increasing attention.
For example, \cite{Liu2021PostTrainingQF} incorporates an innovative ranking loss to preserve the functionality of the self-attention mechanism during quantization, successfully quantizing linear operations (matrix multiplications) in ViTs.
Additionally, FQ-ViT \cite{Lin2021FQViTPQ} further introduces Power-of-Two Factor (PTF) and Log-Int-Softmax (LIS) to quantize the \emph{hardware- and quantization-unfriendly non-linear} operations (i.e., LayerNorm and Softmax) in ViTs,  achieving full quantization.
However, these works are developed for standard ViTs and cannot capture quantization opportunities offered by efficient ViTs \cite{cai2022efficientvit, Han2023FLattenTV}, which feature Softmax-free attention with linear computational complexity to win both quantization accuracy and hardware efficiency.

\vspace{-0.6em}
\subsection{Efficient ViTs}
ViTs \cite{vit, deit, Liu2021SwinTH, cai2022efficientvit, Han2023FLattenTV, Mehta2021MobileViTLG, Graham2021LeViTAV} have gained growing attention recently and have been developed rapidly in the computer vision field. Among them, ViT \cite{vit} firstly applies a pure Transformer to process sequences of image patches, achieving remarkable performance.
Furthermore, DeiT \cite{deit} offers a better training recipe for ViT, significantly reducing training costs.
However, standard ViTs still incur intensive computational costs and huge memory footprints during inference to achieve superior performance, calling for efficient ViTs \cite{Liu2021SwinTH, cai2022efficientvit, Han2023FLattenTV, Mehta2021MobileViTLG, Graham2021LeViTAV}.
Particularly, EfficientViT \cite{cai2022efficientvit}, the SOTA one, replaces the vanilla Softmax-based self-attention of quadratic complexity with a novel Softmax-free lightweight multi-scale attention, achieving a global receptive ﬁeld while enhancing hardware efficiency.
Besides, Flatten Transformer \cite{Han2023FLattenTV} opts for an innovative Softmax-based focused linear attention, preserving expressiveness with low computational complexity.
Despite the inherent algorithmic benefits of Softmax-free linear attentions in efficient ViTs \cite{cai2022efficientvit, Han2023FLattenTV}, including \textbf{(i)} the Softmax-free property to facilitate quantization and \textbf{(ii)} the linear complexity to boost hardware efficiency, their dedicated quantization and acceleration methods remain under-explored.

\vspace{-0.6em}
\subsection{Transformer Accelerators}
Recently, various works \cite{Lu2021SangerAC, Qu2022DOTADA, Li2022AutoViTAccAF, Dass2022ViTALiTyUL, You2022ViTCoDVT, Sun2022VAQFFA} have developed dedicated accelerators to promote Transformers' real-world deployment. Specifically, Sanger \cite{Lu2021SangerAC} dynamically prunes attention maps during inference and builds a reconfigurable accelerator with a score-stationary dataflow to accelerate such sparse patterns.
As for ViT accelerators,
VAQF \cite{Sun2022VAQFFA} accelerates ViTs with binary weights and mixed-precision activations.
Auto-ViT-Acc \cite{Li2022AutoViTAccAF} incorporates heterogeneous computing resources available on FPGAs (i.e., DSPs and LUTs) to separately accelerate the mixed quantization schemes (i.e., fixed-point and power-of-two) for ViTs.
ViTCoD \cite{You2022ViTCoDVT} prunes and polarizes ViTs' attention maps into denser and sparser ones, and constructs an accelerator to execute them on separate computing engines.
While these methods can enhance the hardware efficiency for standard ViTs, they are not directly applicable for efficient ViTs \cite{cai2022efficientvit, Han2023FLattenTV} due to their distinct model architectures, such as Softmax-free linear attention and Convolution-Transformer hybrid structures, calling for dedicated accelerators.

\vspace{-0.6em}
\section{Preliminaries} \label{sec:Preliminaries}

\subsection{Structure of Standard ViTs} \label{sec:arch_evit}
\begin{figure}[t]
	\centerline{\includegraphics[width=0.95\linewidth]{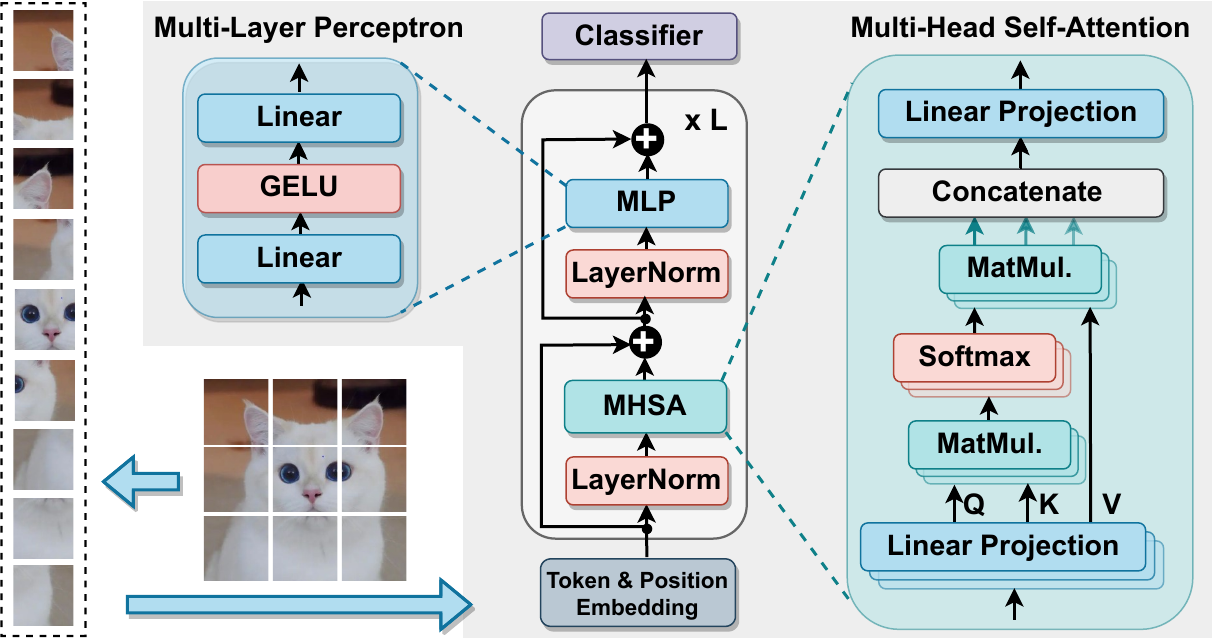}}
	\vspace{-0.6em}
	\caption{The architecture of standard ViTs \cite{vit, deit} that include multiple Transformer blocks. Each block includes an MHSA and an MLP. `MatMul.' is the abbreviation for `Matrix Multiplication'. } 
	\label{fig:vit_arch} \vspace{-0.3em}
\end{figure}

As depicted in Fig. \ref{fig:vit_arch}, input images are initially partitioned into fixed-size patches and further enhanced with token and positional embedding, serving as input tokens for ViTs' Transformer blocks. Among them, each Transformer block comprises two key components: a Multi-Head Self-Attention module (\textbf{MHSA}) and a Multi-Layer Perceptron (\textbf{MLP}), both preceded by LayerNorm (LN) and followed by residual connections. 
Specifically, the \underline{{\textit{MHSA}}} is the core element in Transformers for global information capture. It projects input tokens $X$ into queries $Q_i$, keys $K_i$, and values $V_i$ following Eq. (\ref{eq:qkv_proj}), where $W^Q_i$, $W^K_i$, and $W^V_i$ are corresponding weights for the $i^{th}$ head. 
Subsequently, as formulated in Eq. (\ref{eq:attn}), $Q_i$ is multiplied by the transposed $K_i$ (i.e., $K_i^T$) and then subjected to Softmax normalization (where $d_i$ represents the feature dimension of each head) to generate the attention map. This attention map is further multiplied by $V_i$ to obtain the attention output $\mathbf{A_i}$ for the $i^{th}$ head. Finally, the attention outputs from all $H$ heads are concatenated and projected using weights $W^{O}$ to produce the final MHSA output, i.e., $\mathbf{O_\text{MHSA}}$ in Eq. (\ref{eq:attn_concat}).
As for the \underline{{\textit{MLP}}}, it comprises two linear layers separated by the GELU activation function.
\begin{equation}
    {[Q_i, K_i, V_i]=X\cdot [W^Q_i, W^K_i, W^V_i]}, \label{eq:qkv_proj}
\end{equation} \vspace{-0.8em}
\begin{equation}
    \mathbf{A_i}={\text{Softmax}}(\frac{{Q_iK^T_i}}{\sqrt{d_i}})\cdot {V_i}, \label{eq:attn}
\end{equation} \vspace{-0.8em}
\begin{equation}
    \mathbf{O_\text{MHSA}}=\text{concat}(\mathbf{A_0,A_1,...,A_H})\cdot {W^{O}}. \label{eq:attn_concat}
\end{equation} \vspace{-0.8em}

\textbf{Limitations.} There are two limitations of standard ViTs: 
{\textbf{(i)}} the quadratic computational complexity of the self-attention w.r.t. token numbers \cite{Liu2021SwinTH, cai2022efficientvit, Han2023FLattenTV} and 
{\textbf{(ii)}} the hardware- and quantization-unfriendly non-linear operations, i.e., LN, GELU, and especially Softmax \cite{Lin2021FQViTPQ, Yuan2021PTQ4ViTPQ}, which hinder ViTs' achievable hardware efficiency and quantization accuracy. 

\begin{figure}[t]
	\centerline{\includegraphics[width=\linewidth]{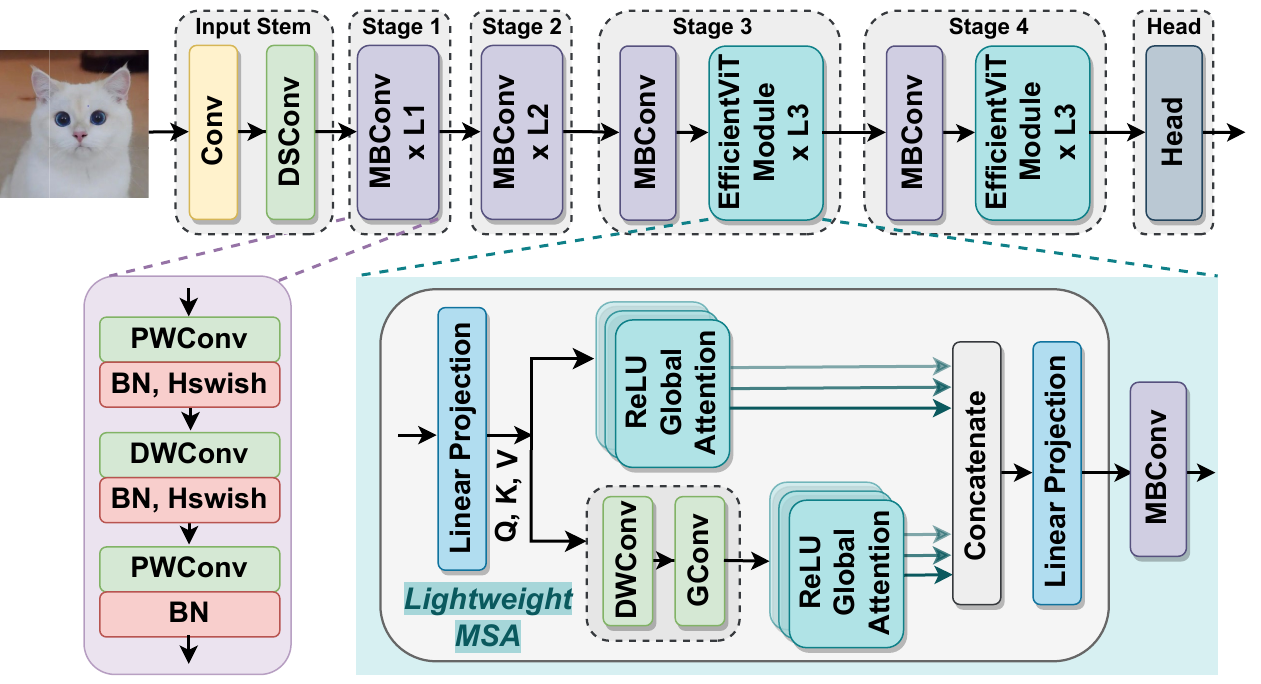}}
	\vspace{-0.6em}
	\caption{The architecture of EfficientViT \cite{cai2022efficientvit} that mainly compromises MBConvs \cite{Sandler2018MobileNetV2IR} and EfficientViT modules.} 
	\label{fig:evit_arch} \vspace{-1.3em}
\end{figure} 

\vspace{-0.6em}
\subsection{Structure of EfficientViT} \label{sec:pre_evit}
To tackle the above limitations, efficient ViTs \cite{cai2022efficientvit, Han2023FLattenTV, Dass2022ViTALiTyUL} have emerged as a promising solution. 
Here we take EfficientViT \cite{cai2022efficientvit}, the SOTA efficient ViT, as the example for illustration. As shown in Fig. \ref{fig:evit_arch}, it not only {\textbf{(i)}} incorporates Softmax-free linear attention but also {\textbf{(ii)}} replaces vanilla LN and GELU with hardware- and quantization-friendly BatchNorm (BN) and Hardswish (Hswish) \cite{Howard2019SearchingFM}, respectively, significantly facilitating quantization and acceleration.

Specifically, EfficientViT \cite{cai2022efficientvit} mainly consists of two types of blocks: \textbf{MBConvs} \cite{Sandler2018MobileNetV2IR} and \textbf{EfficientViT modules}. 
Each \underline{{\textit{MBConv}}} comprises two point-wise convolutions (PWConvs) divided by a depthwise convolution (DWConv). Each layer is followed by a BN and a Hswish (except the last layer). Particularly, BN can be implemented using $1$$\times$$1$ convolutions and seamlessly folded into preceding convolutions, simplifying quantization and acceleration \cite{Zhang2022WSQAdderNetEW}. 
Additionally, each \underline{{\textit{EfficientViT module}}} includes a \textit{lightweight Multi-Scale Attention (MSA)} for context information extraction and an \textit{MBConv} for local information extraction. 
In MSA, inputs are projected to generate $Q/K/V$, which are then processed by lightweight small-kernel convolutions to generate multi-scale tokens. After applying ReLU-based global attention, the results are concatenated and projected to produce final outputs.
Notably, ReLU-based global attention essentially replaces the similarity function $\text{Exp}(QK^T/\sqrt{d})$ in vanilla Softmax-based attention with $\text{ReLU}(Q)\text{ReLU}(K)^T$, thus allowing for \textbf{(i)} the removal of Softmax and \textbf{(ii)} the utilization of the associative property of matrix multiplication to reduce computational complexity from quadratic to linear. This reformulates Eq. (\ref{eq:attn}) as follows:
\begin{equation}
    \mathbf{A_i}=\frac{\text{ReLU}(Q_i)(\sum_{j=1}^{N}\text{ReLU}(K_j)^{T} V_j)}{\text{ReLU}(Q_i)(\sum_{j=1}^{N}\text{ReLU}(K_j)^{T})}. \label{eq:linear_attn}
\end{equation} \vspace{-0.8em}
\section{Trio-ViT's Post-Training Quantization} \label{sec:quantization_alg}
As illustrated above, due to the inherent benefits of the SOTA efficient ViT, dubbed EffcientViT \cite{cai2022efficientvit}, i.e., the vanilla Softmax-based self-attention with quadratic complexity, LN, and GELU, are replaced with hardware- and quantization-friendly ReLU-based global attention with linear complexity, BN, and Hardswish, respectively, we thus explore quantization and acceleration on top of \textbf{EfficientViT} to win both quantization accuracy and hardware efficiency.

We adopt the most widely applied hardware-friendly quantization setting \cite{Li2021BRECQPT} by default, i.e., the \textit{symmetric layer-wise} and \textit{filter-wise uniform quantization} for activations $X$ and weights $W$, respectively. Formally, as expressed in Eq. (\ref{eq:uniform_quantization}), $X_Q$/$W_Q$ are quantized $X$/$W$, $S_a$/$S_w$ are the corresponding scaling factors, $\lfloor \cdot \rceil$ means rounding to the nearest, and $b$ is the quantization bit-width. 
\shh{Particularly, we follow the SOTA PTQ method BRECQ \cite{Li2021BRECQPT}, which uses the {diagonal Fisher Information Matrix (FIM)} to sequentially reconstruct \textit{basic blocks} (e.g., MBConvs and Lightweight MSAs in EfficientViT), thus enhancing cross-layer dependency while maintaining generalizability.
Given the FIM as the objective function, quantization of weights and activations are optimized via Adaround \cite{adaround} and Learned Step size Quantization (LSQ) \cite{Esser2019LearnedSS}, respectively.}
\begin{equation}
\small
    {X_Q} = \text{clip}(\lfloor\frac{{X}}{{S_a}} \rceil,0,2^b-1), \
    {W_Q} = \text{clip}(\lfloor\frac{{W}}{{S_w}} \rceil,0,2^b-1). \label{eq:uniform_quantization}
\end{equation} \vspace{-0.8em}

\vspace{-1.2em}
\subsection{Observations} \label{sec: observations}
\shh{As we adopt the block-wise reconstruction following \cite{Li2021BRECQPT} for quantization optimization} and MBConvs/lightweight MSAs are two primary blocks in EfficientViT, we begin by retaining matrix-multiplications (MatMuls) within MSAs (i.e., the computations in Eq. (\ref{eq:linear_attn})) at full precision to assess quantization impact on MBConvs.

\subsubsection{Observations on Quantization of MBConvs} \label{sec:observe-mbconv}
\begin{figure}[t]
    \centerline{\includegraphics[width=0.9\linewidth]{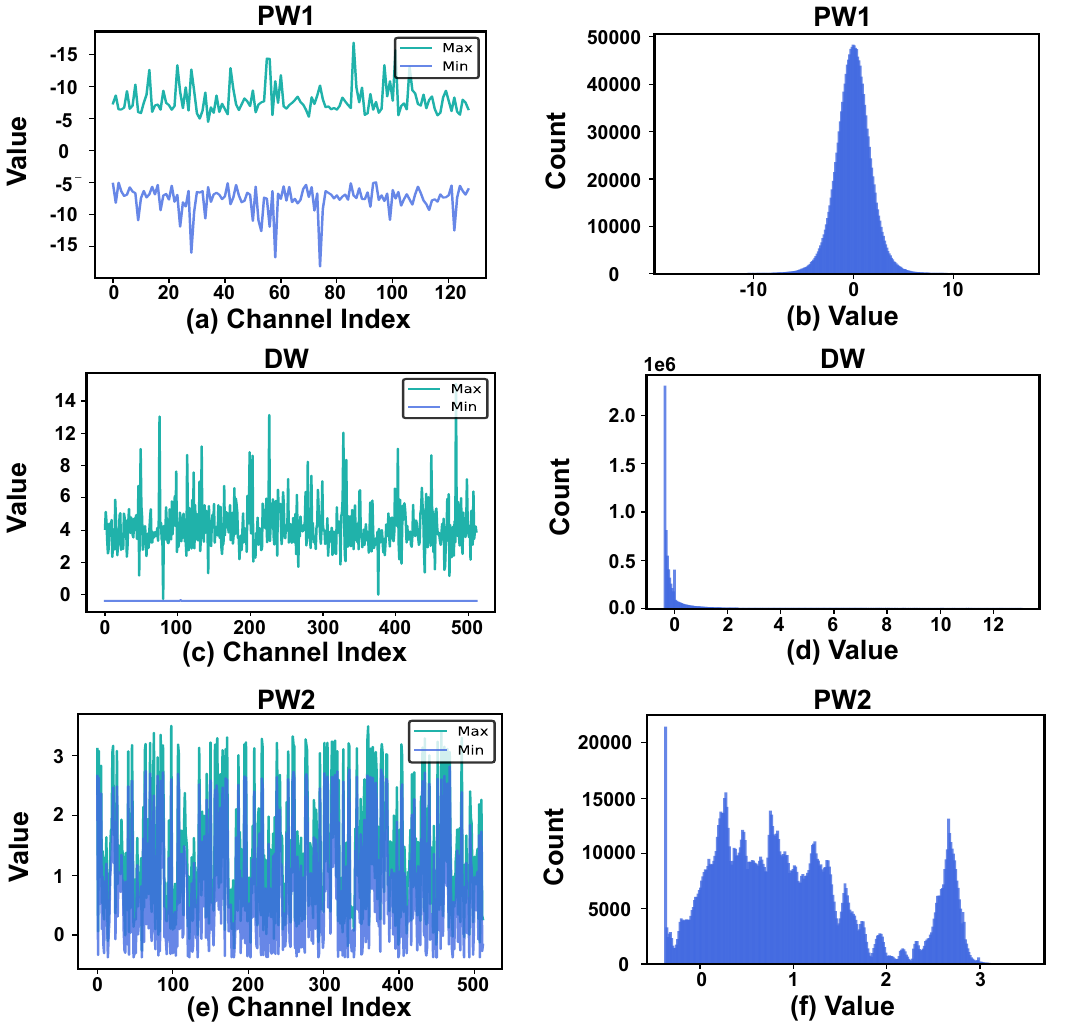}}
	\vspace{-1em}
	\caption{The minimum/maximum values along the channel dimension as well as distributions of input activations in (a) (b) the first PWConv (PW1), (c) (d) DWConv (DW), and (e) (f) the second PWConv (PW2), respectively, within the MBConv in the last stage of EfficientViT-B1 \cite{cai2022efficientvit}.} 
	\label{fig:visul} \vspace{-1.0em}
\end{figure}

Although it has been widely recognized that activations are more sensitive to quantization than weights \cite{Lin2021FQViTPQ, Xiao2022SmoothQuantAA}, this sensitivity is exacerbated in the context of EfficientViT. As presented in Table \ref{tab:w8ax}, quantizing only weights in EfficientViT-B1 \cite{cai2022efficientvit} to $8$-bit (W8) leads to a comparable accuracy ($\uparrow$$0.01\%$) compared to its full-precision counterpart, while quantizing both weights and activations (except MatMuls in MSAs) to the same bit (W8A8) yields a catastrophic accuracy drop of $\downarrow$$\mathbf{76.15\%}$. This emphasizes the extreme quantization sensitivity of activations in EfficientViT, particularly those in MBConvs, as evident in the accuracy comparison between columns $1$ and $2$ in Table \ref{tab:w8a8_ablation}. 
As previously introduced in Sec. \ref{sec:pre_evit}, each MBConv contains two PWConvs separated by a DWConv, we then conduct ablation studies to individually quantize input activations of these three layers in all MBConvs.
As shown in Table \ref{tab:w8a8_ablation}, the input activations of \textbf{DWConvs} (DW) and the \textbf{second} \textbf{PWConvs} (PW2) are the most quantization sensitive and should be primarily responsible for the accuracy drop.
To comprehend this issue, we visualize their input activations in Fig. \ref{fig:visul} and observe two challenges. 
\begin{table}[t]
    \centering
    \setlength{\tabcolsep}{4pt}
    \caption{Top-1 Accuracy of EfficientViT-B1 when weights are all quantized to $8$-bit and activations (except MatMuls in MSAs) are quantized to different bits} \vspace{-0.8em}
    \renewcommand{\arraystretch}{1.2}
    \resizebox{\linewidth}{!}{
    \begin{threeparttable}{
    \begin{tabular}{c|ccccc} \Xhline{3\arrayrulewidth}
         \textbf{EfficientViT-B1\cite{cai2022efficientvit}}  & \textbf{W8}             & \textbf{W8A16}              & \textbf{W8A12}            & \textbf{W8A10} & \cellcolor{n1!30} \textcolor{dark-red}{\textbf{W8A8}} \\ \hline \hline
         \textbf{Top-1 Accuracy* (\%) }             & 79.39          & 79.32              & 78.86   &  75.08        & \cellcolor{n1!30} \textbf{\textcolor{dark-red}{3.23}} \\
         \textbf{Drop (\%)}             & $\uparrow$0.01 & $\downarrow$0.06   & $\downarrow$0.52  & $\downarrow$4.30 & \cellcolor{n1!30} \textcolor{dark-red}{$\downarrow$\textbf{76.15}}\\ \Xhline{3\arrayrulewidth}
    \end{tabular}}
    \begin{tablenotes}
		\footnotesize
		\item[*] Tested on ImageNet with the input size of $224\times224$ by default.
	  \end{tablenotes} 
    \end{threeparttable}} \vspace{-1em}
    \label{tab:w8ax}
\end{table}

\begin{table}[t]
    \centering
    \setlength{\tabcolsep}{2pt}
    \caption{Accuracy of EfficientViT-B1 \cite{cai2022efficientvit} when weights and activations (except MatMuls in MSAs) are both quantized to $8$-bit} \vspace{-0.8em}
    \renewcommand{\arraystretch}{1.2}
    \resizebox{\linewidth}{!}{
    \begin{threeparttable}{
    \begin{tabular}{c|ccccc} \Xhline{3\arrayrulewidth}
         \textbf{EfficientViT-B1}  & \cellcolor{n1!30} \textcolor{dark-red}{\textbf{Head*+MBConvs}}             & \textbf{Head*}        & \textbf{Head*+PW1}      & \cellcolor{n1!30} \textcolor{dark-red}{\textbf{Head*+DW}}            & \cellcolor{n1!30} \textcolor{dark-red}{\textbf{Head*+PW2}}  \\ \hline \hline
         \textbf{Accuracy (\%)}             & \cellcolor{n1!30} \textcolor{dark-red}{\textbf{3.23}}          & 79.24       & {79.13}        & \cellcolor{n1!30} \textcolor{dark-red}{\textbf{28.72}}   &  \cellcolor{n1!30} \textcolor{dark-red}{\textbf{8.85}}        \\
         \textbf{Drop (\%)}             & \cellcolor{n1!30} \textcolor{dark-red}{$\downarrow$\textbf{76.15}} & $\downarrow$0.15  & $\downarrow${0.26}  & \cellcolor{n1!30} \textcolor{dark-red}{$\downarrow$\textbf{50.68}}  & \cellcolor{n1!30} \textcolor{dark-red}{$\downarrow$\textbf{70.55}} \\ \Xhline{3\arrayrulewidth}
    \end{tabular}}
    \begin{tablenotes}
		\footnotesize
		\item[*] ``Stem+Head'' is abbreviated as ``Head'' here for simplicity.
	  \end{tablenotes} 
    \end{threeparttable}} 
    \vspace{-1.0em}
    \label{tab:w8a8_ablation}
\end{table}
\textbf{Challenge \# 1: Inter-Channel Variations in DW's Inputs.} Specifically, as shown in Fig. \ref{fig:visul} (c), the input activation of DW exhibits {significant inter-channel variations}, resulting in the majority of values being represented with few quantization bins (see Fig. \ref{fig:visul} (d)). For example, in the input activation of DW within the MBConv from the last stage in EfficientViT-B1, approximately $\mathbf{90}\%$ values occupy a mere $\mathbf{2.3}\%$ of total quantization bins. In contrast, this percentage is considerably higher at $\mathbf{12.4}\%$ for PW1's input, which is $\mathbf{5.81}\times$ greater.

\textbf{Challenge \# 2: Inter-Channel Asymmetries in PW2's Inputs.}
As depicted in Fig. \ref{fig:visul} (e), input activation of PW2 features extreme inter-channel asymmetries compared to that of PW1 in Fig. \ref{fig:visul} (a), yielding a broader value range and thus a lower quantization resolution. For instance, in PW2's input within the MBConv from the last stage of EfficientViT-B1, the interval of the first channel is ($3.11$, $2.66$), while the interval among all channels is ($3.49$, $-0.38$), which is $\mathbf{8.64}\times$ larger.

\begin{figure}[t]
	\centerline{\includegraphics[width=0.95\linewidth]{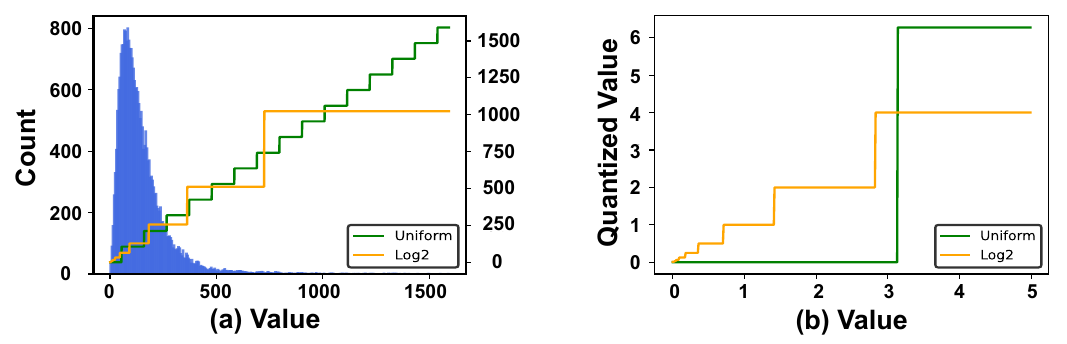}}
	\vspace{-0.8em}
	\caption{(a) gives the distribution of divisor within the MSA from the last stage of EfficientViT-B1 \cite{cai2022efficientvit}, with visualizations of quantization bins of uniform and log2 quantization.  (b) shows quantization bins of small values near zero. } 
	\label{fig:msa_visul} \vspace{-0.5em}
\end{figure}
\subsubsection{Observations on Quantization of Lightweight MSAs} \label{sec:log-observe}
When quantizing MatMuls in Eq. (\ref{eq:linear_attn}) within lightweight MSAs to 8-bit, we encounter notably worse results, which manifest as a ``Not-a-Number" (NaN) issue. We find that this issue primarily arises from the quantization of divisors/denominators in Eq. (\ref{eq:linear_attn}).
As depicted in Figs. \ref{fig:msa_visul} (a) and (b), the wide range of values within divisors results in a reduced quantization resolution for smaller values when adopting the uniform quantization. Nevertheless, \textbf{smaller values within divisors exhibit much greater sensitivity} compared to larger values. For instance, rounding a divisor of $750$ to $1500$ during quantization results in an absolute difference of $\mathbf{750}$, yet it only \textbf{\textit{doubles}} the final division results. In contrast, 
approximating a divisor of $0.01$ to $1$ yields a negligible $\mathbf{0.99}$ absolute difference, but it causes the final results $\mathbf{100}\times$ larger.
These examples distinctly underscore the inherent incompatibility of uniform quantization for divisors, especially those exhibiting a wide range of values.

\begin{figure}[t]
	\centerline{\includegraphics[width=\linewidth]{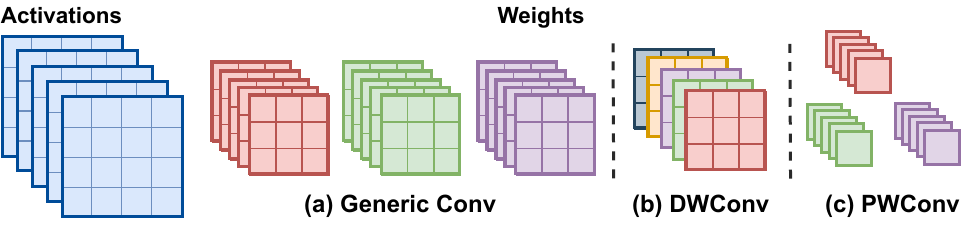}}
	\vspace{-0.8em}
	\caption{Illustrating the layer-wise quantization for activations and the filter-wise quantization for (a) generic convolution (Conv), (b) DWConv, and (c) PWConv. Pixels represented in different colors undergo quantization with distinct scaling factors.} 
	\label{fig:quant_scheme} \vspace{-1.0em}
\end{figure}

\begin{figure}[t]
	\centerline{\includegraphics[width=0.9\linewidth]{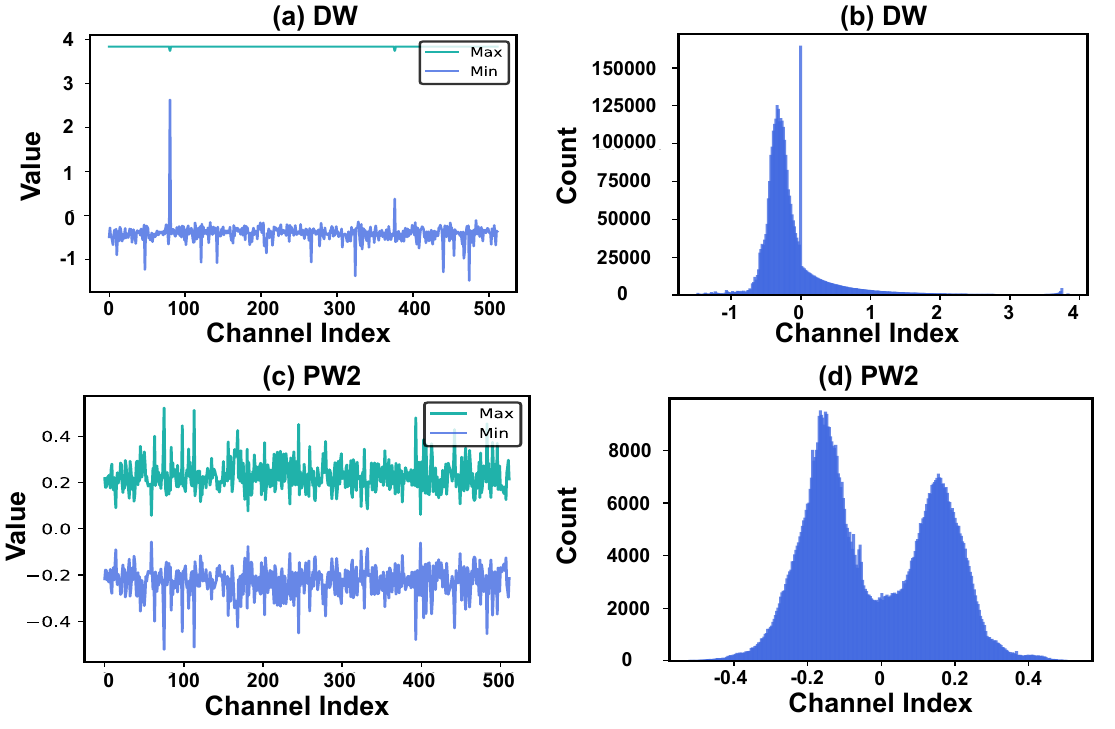}}
	\vspace{-0.8em}
	\caption{(a) and (b) visualize the input of DWConv (DW) after channel-wise migration, and (c) and (d) draw the input of the second PWConv (PW2) after filter-wise shifting. We take the MBConv in the last stage of EfficientViT-B1 \cite{cai2022efficientvit} as the example.} 
	\label{fig:shift_scaled_dis} \vspace{-1.5em}
\end{figure}
\vspace{-0.6em}
\subsection{Channel-Wise Migration for DW's Inputs} \label{sec: cw-qaunt}
As introduced in \textbf{Challenge \# 1} in Sec. \ref{sec:observe-mbconv}, there exist extreme inter-channel variations in DW's inputs, making the vanilla layer-wise quantization unsuitable. 
Fortunately, owing to the distinctive algorithmic characteristic of DW, which processes each input channel independently and thus eliminates the summations along different channels, we can directly employ \textit{\textbf{channel-wise quantization}} to assign individual scaling factors for each input channel. This approach effectively addresses the above challenge without compromising hardware efficiency.
Nonetheless, despite its potential advantages in enhancing quantization accuracy, it significantly increases the number of scaling factors for activations, posing another challenge for their optimizations via LSQ \cite{Esser2019LearnedSS}, which is widely adopted to optimize activation quantization.

To address this limitation, we propose to adopt \textit{\textbf{channel-wise migration} on top of the \textbf{layer-wise quantization}} for DW's inputs.
Specifically, as shown in Fig. \ref{fig:quant_scheme}, weights in DW feature the same channel number as inputs, and each weight channel functions as an independent filter dedicated to processing its corresponding input channel. This arrangement enables us to assign distinct scaling factors to each weight channel. Thus, as depicted in Fig. \ref{fig:quant_scheme} (b), filter-wise quantization is essentially the channel-wise quantization for DW's weights.
Consequently, inter-channel variations of activations $A$ can be seamlessly transferred to weights $W$ through a mathematically equivalent transformation employing channel-wise migration factors $M_i$:  
\vspace{-0.3em}
\begin{equation} 
    O^i = \frac{A^i}{M^i}\times (W^iM^i) = S_a\cdot Q(\frac{A^i}{M^i})\times S_w^i\cdot Q(W^iM^i), \label{eq:channel-scaling} \vspace{-0.3em}
\end{equation} 
where $O_i, A_i$, $W_i$, and $M_i$ are the output, input, weight, and migration factor for $i^{th}$ channel, respectively. $Q(\cdot)$ donate the quantization function, $S_a$ and $S_w^i$ are the layer-wise and channel-wise scaling factors for $A$ and $W_i$, respectively. This approach greatly facilitates activation quantization without increasing learnable scaling factors of activations and impeding weight quantization. Note that weights can be pre-transformed before deployment to eliminate the on-chip computation. As for activations that depend on the input images during inference and thus cannot be pre-processed, we can fuse $M^i$ with $S_a$ to obtain a fused scaling factor $S_{am}^i$ in advance, thus avoiding the on-the-fly transformation: 
\begin{equation}
    Q(\frac{A^i}{M^i}|\ S_a) = \frac{1}{S_a}\cdot\frac{A^i}{M^i} = \frac{A^i}{S^i_{am}} = Q(A_i|\ S_{am}^i).
\end{equation}
Furthermore, the computation of $M_i$ can be formulated as: \vspace{-0.5em}
\begin{equation}
    \text{mean}(A)=\frac{1}{N}\sum_{i=1}^{N}\text{max}(A_i), \ M_i=\frac{\text{max}(A_i)}{\text{mean}(A)}, \label{eq:migration} \vspace{-0.5em}
\end{equation}
where $\text{mean}(A)$ is the mean of the maximal values across $N$ channels. By comparing Figs. \ref{fig:visul} (c)/(d) and Figs. \ref{fig:shift_scaled_dis} (a)/(b), it is evident that this approach can accomplish two essential objectives. \underline{{\textit{Firstly}}}, it compresses outliers, effectively reducing the value range of activations. \underline{{\textit{Secondly}}}, it amplifies smaller values, making them more amenable to quantization.

\vspace{-0.6em}
\subsection{Filter-Wise Shifting for PW2's Inputs} \label{sec: fw-qaunt}
To eliminate the inter-channel asymmetries in PW2’s inputs (introduced in \textbf{Challenge \# 2} of Sec. \ref{sec:observe-mbconv}), inspired by \cite{Wei2023OutlierSA}, we propose filter-wise shifting to pre-process PW2’s inputs before quantization.
As expressed in Eq. (\ref{eq:channel-shift}), we subtract each input channel by its channel-wise mean $c^i$, thus obtaining the calibrated input $\hat{A}$ centered around zero (see Fig. \ref{fig:shift_scaled_dis} (c)) and significantly reducing value ranges (see the comparison between Figs. \ref{fig:visul} (f) and \ref{fig:shift_scaled_dis} (d)). 
To accommodate the above filter-wise shifting for activations and keep the same functionality as the original PW, the original bias $b^j$ of the $i^{th}$ output channel needs to be updated to $\hat{b^j}$ following Eq. (\ref{eq:shift-bais}), where $N$ donate the input channel number. This bias update can be pre-computed to eliminate the on-chip processing. 
\begin{equation}
    c^i = \frac{\text{max}(A^i)-\text{min}(A^i)}{2}, \ \hat{A^i} = A^i - c^i. \label{eq:channel-shift} \vspace{-0.5em}
\end{equation} 
\begin{equation}
\small
    O^j = A\cdot W^j + b^j = \hat{A}\cdot W^j + (\sum_{i=1}^N c^i w^{(i,j)} + b^j) =  \hat{A}\cdot W^j + \hat{b^j}. \label{eq:shift-bais}
\end{equation} \vspace{-0.8em}

\vspace{-0.6em}
\subsection{Log2 Quantization for Divisors in MSAs} \label{sec: log2-qaunt}
As illustrated in Figs. \ref{fig:msa_visul} (a) and (b), log2 quantization will allocate more quantization bins to smaller values and vice versa. This inherent characteristic aligns with the algorithmic property of divisors in MSAs, where small values exhibit higher quantization sensitivity as explained in Sec. \ref{sec:log-observe}. Thus, to improve the quantization resolution of small values in divisors of MSAs, we advocate adopting log2 quantization:
\begin{equation}
\begin{aligned}
    &X_Q^{log2} = \text{clip}(\lfloor\text{log}_2{(S_{Q^R} S_{K_{sum}^R}X_Q)}\rceil, -a, b) \\
    &= \text{clip}(\lfloor\text{log}_2{(S_{Q^R} S_{K_{sum}^R})\rceil+ \lfloor\text{log}_2(X_Q)}\rceil, -a, b),
\end{aligned}
\end{equation}
where $X_Q^{log2}$ is log2-qauntized divisors, $X_Q$ is the integer divisors generated by integer multiplications between $\text{ReLU}(Q)$ and $\sum_{j=1}^{N}\text{ReLU}(K_j)^{T}$ in Eq. (\ref{eq:linear_attn}), and $S_{Q^R}$ and $S_{K_{sum}^R}$ are their scaling factors. Note that $\lfloor\text{log}_2{(S_{Q^R} S_{K_{sum}^R})}\rceil$ can be pre-computed, while $\lfloor\text{log}_2(X_Q)\rceil$ can be effectively implemented in the integer domain following \cite{Lin2021FQViTPQ}. Specifically, we first adopt the leading one detector (LOD) to find the index $i$ of the first non-zero bit of $X_Q$, then add $i$ with the value of the $(i\mbox{-}1)^{th}$ bit to obtain the result.
For instance, if $X_Q$ is ${(0110\ 0011)}_2$, then the index $i$ of the first non-zero bit is $6$ and the value of the $5^{th}$ bit is $1$, thus the log2-quantized $X_Q$ ($X_Q^{log2}$) is $7$.

By adopting log2 quantization for divisors, we can further replace hardware-unfriendly divisions in Eq. (\ref{eq:linear_attn}) with hardware-efficient bit-wise shifts, further enhancing hardware efficiency while boosting quantization performance.
\section{Trio-ViT's Accelerator}
\label{sec:Accelerator}

\subsection{Design Considerations} \label{sec:design_consider}
To fully unleash our algorithmic benefits, developing a dedicated accelerator for quantized EfficientViT \cite{cai2022efficientvit} is highly desired. However, this poses several challenges due to \textbf{(i)} various operation types within EfficientViT and \textbf{(ii)} the distinct computational pattern of its lightweight attention compared to the vanilla self-attention in standard ViTs \cite{vit, deit}.

\subsubsection{Design Challenge \# 1: Various Operation Types} \label{sec:design_consider_1}
As shown in Fig. \ref{fig:evit_arch} and introduced in Sec. \ref{sec:arch_evit}, there are mainly four types of operations in the Convolution-Transformer hybrid backbone of EfficientViT: generic convolutions (where output pixels are produced by the accumulation of partial sums within sliding windows along input channels), PWConvs (which essentially are generic convolutions with $1\times 1$ kernels), DWConvs (which process each input channel separately, thus only partial sums within the sliding window need to be accumulated), and matrix multiplications (MatMuls). 
For example, they account for $1.1\%$, $91.9\%$, $5.4\%$, and $1.6\%$ of the total operation numbers in EfficientViT-B1, respectively, when input resolution is $224$$\times$$224$.

\begin{figure*}[t]
	\centerline{\includegraphics[width=\linewidth]{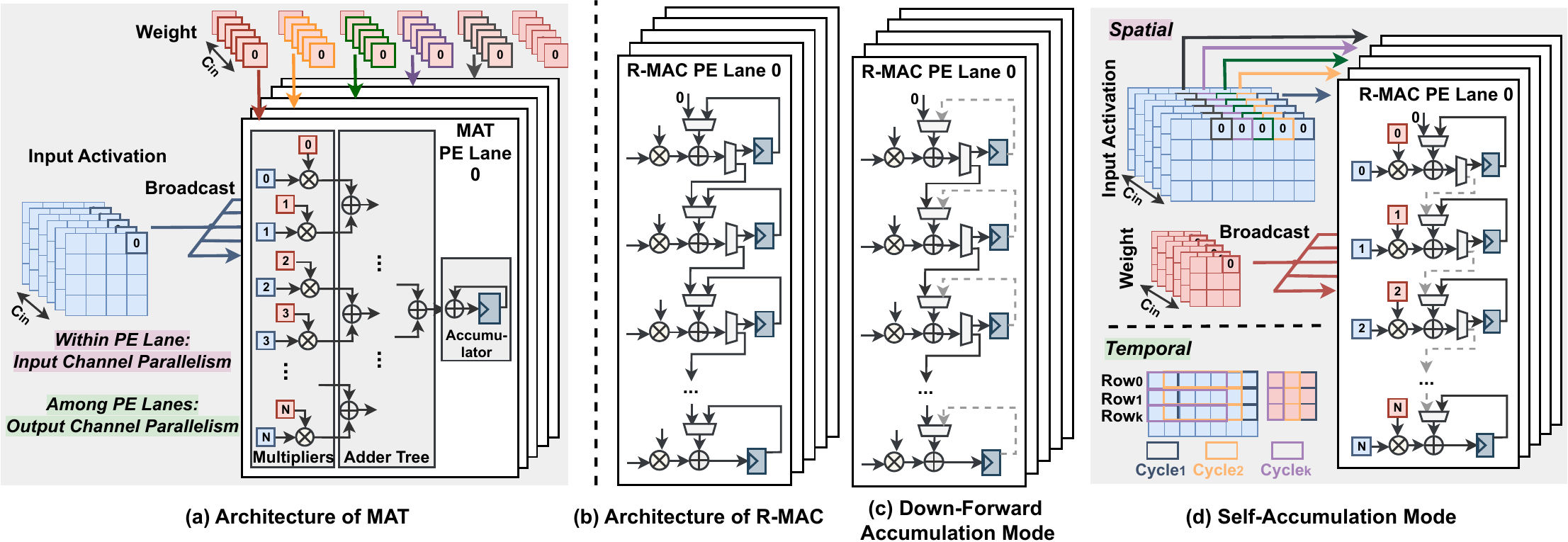}}
	\vspace{-0.6em}
	\caption{(a) illustrates the execution of PWConvs on the MAT architecture. (b) shows the architecture of the reconfigurable design (R-MAC), (c) and (d) depict its down-forward accumulation mode and self-accumulation mode, respectively. } 
	\label{fig:mat_arch} \vspace{-1.0em}
\end{figure*} 

\textbf{Design Choice \# 1: Multipliers-Adder-Tree Architecture.}
Considering that PWConvs are the dominant type of operations, one natural thought is to adopt the Multipliers-Adder-Tree (\textbf{MAT}) architecture, which is a typical design to efficiently support PWConvs with channel parallelism \cite{Yu2020LightOPUAF, Liu2021TowardFA}. 
Specifically, as illustrated in Fig. \ref{fig:mat_arch} (a), each processing element (PE) lane in the MAT engine is responsible for the multiplication (via multipliers), summation (via the adder tree), and accumulation (via the accumulator) along the input channel dimension to generate each output pixel (for PWConvs) or partial sum (dubbed psum, for generic convolutions). Thus, the parallelism \underline{\textit{within the PE lane}} of the MAT engine is along the \textit{input channel dimension} to facilitate \textit{psum reuse}.
Besides, inputs are broadcast to different PE lanes and multiplied with different weight filters to produce output/psum pixels from different output channels, thus the parallelism \underline{\textit{among PE lanes}} is along the \textit{output channel dimension} to enhance \textit{input reuse}.  

\textbf{Limitations.}
Although the MAT architecture can efficiently process PWConvs as well as easily support generic convolutions and MatMuls (which can be treated as PWConvs with large batch sizes), it has limited flexibility when handling DWConvs.
\underline{{\textit{Firstly}}}, as for DWConvs, only psums generated from the same sliding window can be summed and accumulated, thus the achievable parallelism of each PE lane in MAT is limited by kernel sizes of DWConvs.
{\underline{{\textit{Secondly}}}, DWConvs in EfficientViT feature various kernel sizes ($3$$\times$$3$ and $5$$\times$$5$) and strides ($1$ and $2$), resulting in different sizes of sliding windows and distinct overlap patterns between adjacent sliding windows when conducting convolutions. This necessitates extra line buffers and substantial memory management overheads to support the multiplication and summation functionality within PE lanes for generating consecutive output pixels \cite{Yu2020LightOPUAF}. } 
\underline{{\textit{Thirdly}}}, the lack of input reuse opportunities within DWConvs will lead to either a high memory bandwidth requirement or low PE utilization when accommodating the output channel parallelism among PE lanes in the MAT architecture.

\textbf{Design Choice \# 2: Reconfigurable Architecture.}
To solve the above limitations, the {reconfigurable architecture} depicted in Fig. \ref{fig:mat_arch} (b) can be considered, which incorporates multiple Reconfigurable Multiplier-ACcumulation units (\textbf{R-MACs}).
{\textbf{(i)}} As depicted in Fig. \ref{fig:mat_arch} (c), when executing generic convolutions, PWConvs, and MatMuls, this architecture can be configured to operate in the \textit{down-forward accumulation} mode to achieve the same functionality as the MAT architecture. This means that each PE lane supports the input channel parallelism to achieve psum reuse, while different PE lanes facilitate output channel parallelism to exploit input reuse.
{\textbf{(ii)}} As depicted in Fig. \ref{fig:mat_arch} (d), when executing DWConvs, this architecture can be configured to run in the \textit{self-accumulation} mode, thus partial sums within each sliding window can be \textit{temporally} accumulated with each R-MAC, making it inherently supports DWConvs with various kernel sizes.

Specifically, as illustrated Fig. \ref{fig:mat_arch} (d) right, different R-MACs \underline{\textit{within each PE lane}} can \textit{spatially} compute multiple output pixels from different output channels, thus the parallelism is along the \textit{output channel} here.
Besides, as shown in Fig. \ref{fig:mat_arch} (d) top left, weights can be broadcast to all PE lanes and multiplied with input pixels from different sliding windows to generate consecutive output pixels from the same output channel. By doing this, we can reuse overlaps among adjacent sliding windows with only several auxiliary registers and easily support DWConvs with different strides.
For example, as shown in Fig. \ref{fig:mat_arch} (d) bottom left, where we take computations of the $K\times K$ DWConv with a stride of $1$ executed on $M$ R-MACs arranged in the same row from $M$ PE lanes as the example. Initially, a sequence of input pixels $\{a_0, a_1, ..., a_{M-1}, ..., a_{M+K-2}\}$ is transmitted to input shift registers, which is then moved forward by cycles. During each cycle, the first $M$ pixels in shift registers are independently multiplied with the broadcast weight $w_i$, generating the $i^{th}$ psums for $M$ consecutive output pixels. 
After $K$ cycles, the computation moves to the next row of the input feature map and the corresponding filter, continuing in the same computation mode. This process is repeated until all $K$ rows are processed, resulting in $M$ output pixels.
As for computations of DWConvs with a stride of $2$, overlaps among adjacent sliding windows are spaced instead of successive. Thus, odd-column-indexed input pixels within each row are initially transmitted to shift registers for processing, followed by the even-column-indexed pixels. Weights also need to be broadcast following the same ``first odd, then even" rule to accommodate this modified computation scheme. 
Thereby, the parallelism \underline{\textit{among PE lanes}} in this architecture is along the \textit{output feature map} to enhance \textit{weight reuse}.

\textbf{Limitations.}
Despite its flexibility in supporting all types of operations in EfficientViT, there exist reconfigurable overheads in two aspects. 
\underline{{\textit{Firstly}}}, the overheads in computational resources and buffers: each R-MAC needs a high-bit adder and a psum register to support the self-accumulation.
\underline{{\textit{Secondly}}}, the overheads in control logic: extra multiplexers are required to simultaneously support both two accumulation patterns, i.e., self-accumulation and down-forward accumulation.

\textbf{Our Proposed Design: Hybrid Architecture.}
Considering the fact that: {\textbf{(i)}} PWConvs are the dominant operations in EfficientViT \cite{cai2022efficientvit} and the MAT architecture can efficiently support them, as well as {\textbf{{(ii)}}} EfficientViT incorporates various operation types, especially DWConvs with various kernel sizes and strides, and the R-MAC design can flexibly support all of them, we propose a hybrid architecture for our dedicated accelerator to marry the best of both designs.
Specifically, it consists of a \textit{MAT} \textit{engine} to efficiently process generic convolutions, PWConvs, and MatMuls, and a \textit{R-MAC engine} to effectively support the above three operation types and DWConvs, thus enhancing flexibility while maintaining hardware efficiency. 

\textbf{Offered Opportunity: Inter-Layer Pipeline.}
Besides the efficiency and flexibility of our proposed hybrid architecture, it offers an opportunity for inter-layer pipeline processing, thus saving data access costs.
Specifically, DWConvs are exclusively executed on the R-MAC engine in our hybrid accelerator and are sandwiched by two PWConvs in MBConvs of EfficientViT.
Thus, when the R-MAC engine handles DWConvs, the resulting outputs can be immediately transmitted to the idle MAT engine and serve as inputs to participate in the computation of the subsequent PWConvs.
This integration enables the computation of DWConvs and their following PWConvs to be fused, leading to enhanced hardware utilization and reduced data access costs from off-chip.

\subsubsection{Design Challenge \# 2: Distinct Computational Pattern of Attention} \label{sec:hw_challenge2}
As expressed in Eq. (\ref{eq:linear_attn}), after query $Q$ and key $K$ undergo ReLU, there are \textbf{\textit{five}} remaining steps for producing the final attention map $\mathbf{A}$:
\textbf{(i)} MatMuls between $\text{ReLU}(K^T)$ and value $V$;
\textbf{(ii)} toke-wise summation of $\text{ReLU}(K^T)$; then
\textbf{(iii)} MatMuls of $\text{ReLU}(Q)$ and outputs from step i;
\textbf{(iv)} matrix-vector multiplications of $\text{ReLU}(Q)$ and outputs from step ii; and finally
\textbf{(v)} divisions between outputs from step iii (divisors) and step iv (dividends). Thanks to our {log2 quantization} for divisors as introduced in Sec. \ref{sec: log2-qaunt}, the costly divisions can be substituted by hardware-efficient bit-wise shifts. Thus, it is evident that, besides multiplications, there are also element-wise summations and bit-wise shifts involved in computations of lightweight attention in EfficientViT. These multiplication-free element-wise operations are inherently incompatible with our multiplication-based PE arrays. Besides, they exhibit low computational intensity, yielding increased bandwidth requirements or potential delays \cite{Kim2023FullSO}.

\textbf{Our Proposed Solution: Low-Cost Auxiliary Processors.}
In addition to the MAT engine and R-MAC engine, which can be leveraged to effectively process MatMuls in the above steps i, iii, and iv, we further integrate several low-cost auxiliary processors into our hybrid architecture to facilitate the multiplication-free computations involved in lightweight attention.
Particularly, we incorporate an \underline{\textit{adder tree}} to support the row-wise summation in step ii and a \underline{\textit{shifter array}} to handle the bit-wise shifts in step v.
This architectural adjustment offers an opportunity for computation fusion within the attention (i.e., intra-layer pipeline, which will be explained in Sec. \ref{sec:hw_pipeline}), thus enhancing data reuse and easing bandwidth requirements. 

\textbf{Offered Opportunity: Intra-Layer Pipeline.}
Considering the MAT engine, R-MAC engine, and low-cost auxiliary processors in our hybrid design, the above computation steps in the attention that involve various operation types can be simultaneously handled by distinct computing units, thus offering the fusion opportunity.
For example, {\textbf{(i)}} when $\text{ReLU}(K^T)\times V$ in step i are executed on the MAT/R-MAT engine, $\text{ReLU}(K^T)$ can be broadcast to the auxiliary adder tree for performing the row-wise summation in step ii.
Besides, {\textbf{(ii)}} when steps iii and iv are processed, their outputs can be immediately sent to the shifter array for element-wise divisions.

\begin{figure}[t]
	\centerline{\includegraphics[width=\linewidth]{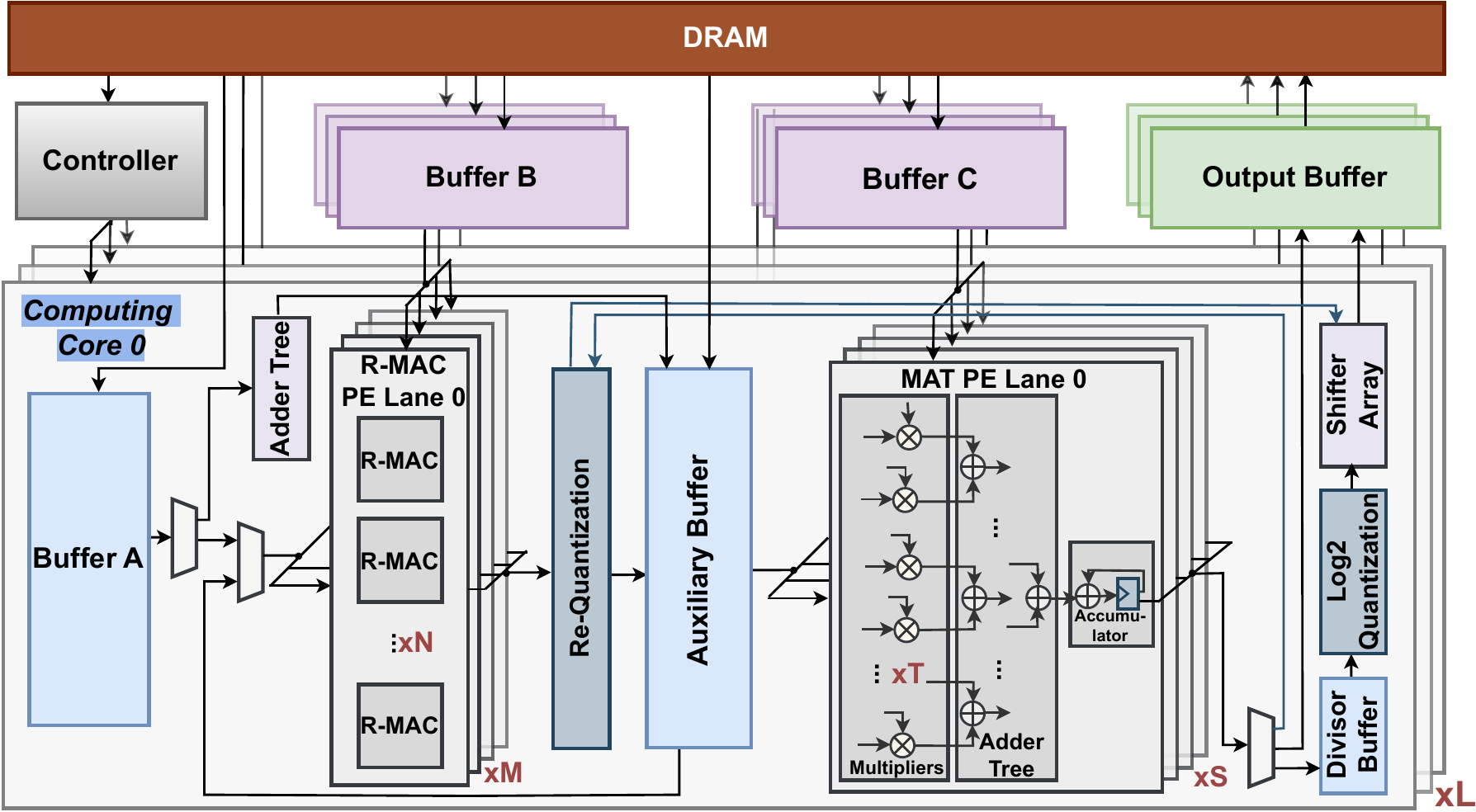}}
	\vspace{-0.3em}
	\caption{Micro-architecture of our accelerator that includes buffers (buffer A/B/C and auxiliary/divisor/output buffers) and computing units (such as R-MAC/MAT engines, auxiliary processors, and re-/log2 quantization modules).} 
	\label{fig:hw_arch} \vspace{-1.2em}
\end{figure} 
\vspace{-0.6em}
\subsection{Micro Architecture} \label{sec:hw-arch}
As shown in Fig. \ref{fig:hw_arch}, our dedicated accelerator consists of $L$ computing cores and several global buffers (buffer B/C and output buffer). Each computing core includes several internal buffers (buffer A and auxiliary/divisor buffers) and multiple computing units (R-MAC engine, MAT engine, auxiliary processors, and re-/log2 quantization modules).
Specifically, as for computing units in each computing core, the \underline{\textit{R-MAC engine}} comprises $M$ PE lanes, each containing $N$ R-MACs and can be reconfigured to operate in either self-accumulation or down-forward accumulation modes. This flexibility enables the effective processing of all multiplication-based operations in EfficientViT, including generic convolutions, DWConvs, PWConvs, and MatMuls, as introduced in \textbf{Design Choice \# 2} in Sec. \ref{sec:design_consider_1}.
The \underline{\textit{MAT engine}} consists of $S$ PE lanes, each including $T$ multipliers, and is developed to efficiently handle multiple multiplication-based operations in EfficientViT, excluding DWConvs, as explained in \textbf{Design Choice \# 1} in Sec. \ref{sec:design_consider_1}.
Besides, {the \underline{\textit{log2 quantization module}} is developed to quantize divisors in Eq. (\ref{eq:linear_attn})} following steps outlined at the end of the first paragraph in Sec. \ref{sec: log2-qaunt}, thus boosting quantization accuracy as well as enabling the conversion of costly divisions into hardware-efficient bit-wise shifts.
Our accelerator is also equipped with several low-cost \underline{\textit{auxiliary processors}}, such as the adder tree and shifter array, to accommodate the computation of multiplication-free operations (e.g., row-wise summation and bit-wise shift) in MSAs, as discussed in \textbf{Our Proposed Solution} in Sec. \ref{sec:hw_challenge2}. 
Additionally, a \underline{\textit{re-quantization module}} is also incorporated to re-quantize outputs following {Eq. (\ref{eq:re-quant})}, where $O_Q, A_Q,$ and $W_Q$ are quantized output $O$, input, and weight, respectively, $S_o, S_a,$ and $S_w$ are their corresponding scaling factors, and $b$/$c$ are both positive integers. By doing this, the floating-point re-scaling factors are converted into dyadic numbers, allowing the re-quantization process to be implemented using integer-only multiplications and bit-wise shifts \cite{Li2022IViTIQ, Yao2020HAWQV3DN, Jacob2017QuantizationAT}, thus facilitating both intra- and inter-layer pipelines on-chip. 
\begin{equation}
    O_Q = \frac{O}{S_o} = \frac{S_aS_w\cdot A_QW_Q}{S_o},\ \text{DN}(\frac{S_aS_w}{S_o}) = \frac{b}{2^c}. \label{eq:re-quant}
\end{equation}

As for internal buffers, buffer A broadcasts data to all PE lanes in the R-MAC engine and can also send data to the auxiliary adder tree. The auxiliary buffer cashes outputs from the R-MAC engine and broadcasts data to all PE lanes in the MAT engine, serving as a bridge between the two engines.
The divisor buffer stores divisors in Eq. (\ref{eq:linear_attn}) and then transfers them to the log2 quantization module, in preparation for the following bit-wise shifts.
Regarding global buffers, buffers B/C send data to all computing cores, where data are distributed and transmitted to different PE lanes in R-MAC/MAT engines. The output buffer stores data from all computing cores and directs them to the off-chip DRAM. 

\begin{figure}[t]	
\centerline{\includegraphics[width=\linewidth]{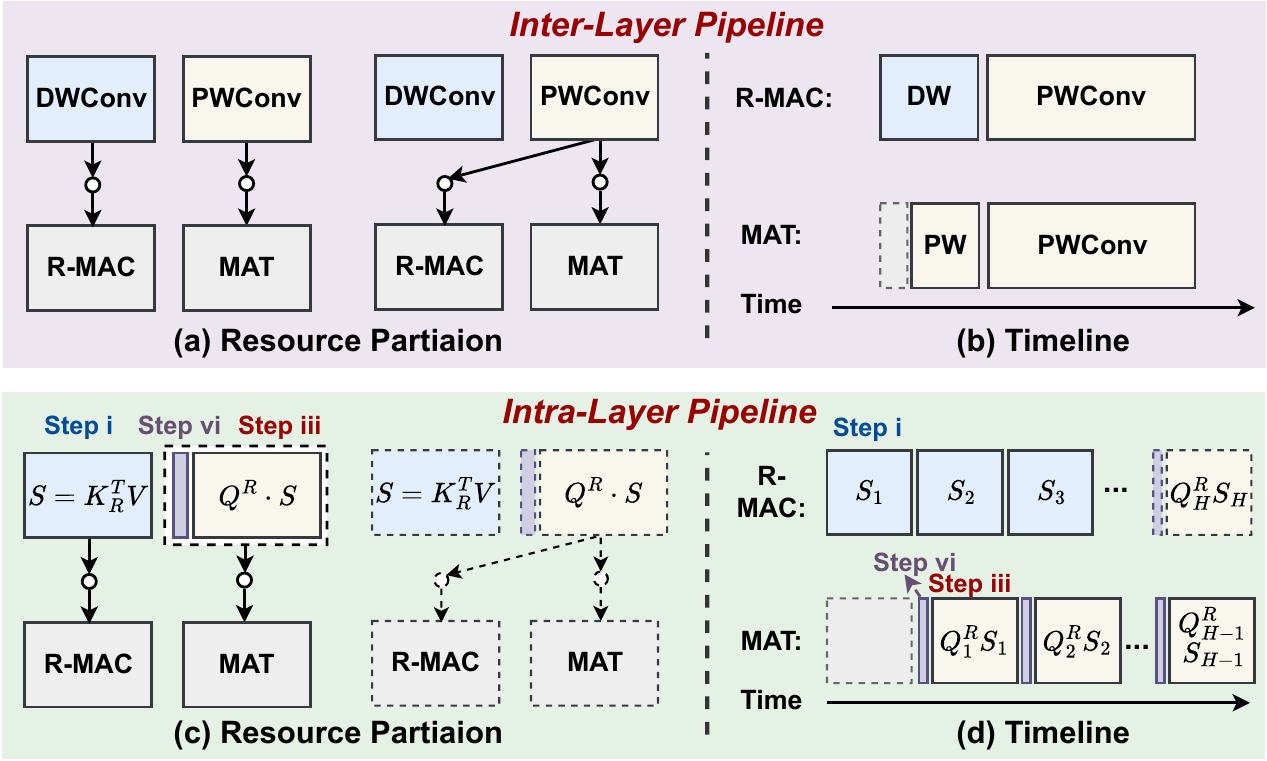}}
\caption{(a)/(b) illustrate the inter-layer pipeline, while (c)/(d) show the intra-layer pipeline. $K_R^T$ and $Q^R$ donates $\text{ReLU}(K)^T$ and $\text{ReLU}(Q)$, respectively.} 
	\label{fig:pipeline} \vspace{-1.2em}
\end{figure} 
\vspace{-0.6em}
\subsection{Inter- and Intra-Layer Pipelines} \label{sec:hw_pipeline}
As introduced above, our dedicated accelerator incorporates both multiplication-based engines (R-MAC and MAT engines, with DWConvs limited to the former) and multiplication-free engines (auxiliary processors that are designed to expedite computations in MSAs). This architecture inherently offers opportunities for pipeline processing, where various operations can be simultaneously executed on distinct computing units, thereby enhancing hardware utilization and throughput.
As for the \underline{\textit{inter-layer pipeline}}, as shown in Figs. \ref{fig:pipeline} (a) and (b), when the R-MAC engine handles DWConv, the resulting outputs are first subtracted by the channel-wise mean obtained on the calibration data to implement the filter-wise shifting introduced in Sec. \ref{sec: fw-qaunt}, aiming to facilitate activation quantization. After that, they undergo re-quantization through the re-quantization module before being stored in the auxiliary buffer. Then, they are promptly directed to the idle MAT engine to serve as inputs for subsequent PWConv computations.
Given that DWConvs entail much fewer computations compared to PWConvs, once the processing of the current DWConv is completed, the R-MAC engine can be reassigned to participate in the concurrent computation of PWConv, alongside the MAT engine.

Regarding the \underline{\textit{intra-layer pipeline}}, {\textbf{(i)}} when the R-MAC engine processes $S_i=\text{ReLU}(K_i)^T\cdot V_i$ for the $i^{th}$ head, $\text{ReLU}(K_i)^T$ are broadcast to the auxiliary adder tree to generate the vector $\text{ReLU}(K_i)^T_\text{sum}$ via row-wise summations. This implies that steps i/ii in the Sec. \ref{sec:hw_challenge2} can be executed simultaneously on distinct computing units.
Concurrently, {\textbf{(ii)}} the MAT engine sequentially preform multiplications between $\text{ReLU}(Q_{i-1})$ and the already obtained $\text{ReLU}(K_{i-1})^T_\text{sum}$ as well as $S_{i-1}$ to generate divisors and dividends in Eq. (\ref{eq:linear_attn}) for the $(i-1)^{th}$ head, respectively. This means that steps iv/iii in the Sec. \ref{sec:hw_challenge2} are computed consecutively on the MAT engine.
During this process, the firstly generated divisors are cached in the divisor buffer and then routed to the log2 quantization module for log2 quantization.
{\textbf{(iii)}} Once the dividends are obtained, they can be re-quantized via the re-quantization module and then sent to the auxiliary shifter array along with the already log2-quantized divisors to conduct element-wise divisions via bit-wise shifts, thus obtaining the final outputs of MSA.
Note that once the computation for $S$ of all heads is finished, the R-MAC engine can be reused to compute divisors and dividends, together with the MAT engine.

\vspace{-0.5em}
\section{Experimental Results}
\label{sec:experiments}

\begin{table}[]
\centering
\setlength{\tabcolsep}{4pt}
\caption{Resource consumption of our dedicated accelerator} \vspace{-0.8em}
\renewcommand{\arraystretch}{1.1}
\resizebox{\linewidth}{!}{
\begin{tabular}{c|cccc}
\Xhline{3\arrayrulewidth}
\textbf{Resources} & \textbf{BRAM} & \textbf{DSP}  & \textbf{LUT}    & \textbf{FF}     \\ \hline \hline
\textbf{Available} & 912           & 2520          & 274080          & 548160          \\
\rowcolor{dark-green!16} \textbf{Used}      & 162 (17.8\%)  & 1024 (40.6\%) & 130628 (47.7\%) & 152819 (27.9\%) \\ \Xhline{3\arrayrulewidth}
\end{tabular}} \label{tab:hw_resources} \vspace{-1.7em}
\end{table}
\subsection{Experimental Setup}
\label{sec:algo-set-up}
\textbf{Dataset, Baselines, and Metrics.}
We validate our Trio-ViT's post-training quantization algorithm on the \textbf{\textit{ImageNet dataset}} \cite{Deng2009ImageNetAL} \shh{and implement it on the NVIDIA GeForce RTX 3090 GPUs, each with $24$GB of memory}. Specifically, we randomly sample $1024$ images from the training set as calibration data and then test on the validation set. 
\ding{182} To verify the effectiveness of our \textbf{\textit{quantization engine}}, we consider \underline{\textbf{\textit{seven baselines}}}:
MinMax, EMA \cite{EMA}, Percentile \cite{Percentile}, OMSE \cite{omse}, Bit-Split \cite{bitsplit}, EasyQuant \cite{Wu2020EasyQuantPQ}, and the SOTA method FQ-ViT \cite{Lin2021FQViTPQ}, for standard ViTs\cite{vit}/DeiTs\cite{deit}, and compare with them in terms of top-1 accuracy.
\ding{183} To validate our dedicated \textbf{\textit{accelerator}}, \shh{we consider \underline{\textbf{\textit{ten baselines}}}: (i) full-precision ViTs executed on the widely-used Edge GPU (NVIDIA Tegra X2);
$8$-bit ViTs quantized following (ii) FasterTransformer \cite{fastertransformer}, (iii) I-BERT \cite{Kim2021IBERTIB}, and (iv) I-ViT \cite{Li2022IViTIQ}, and accelerated on the Turing Tensor Core of NVIDIA 2080Ti GPU, which supports efficient integer arithmetic via TVM; 
and full-precision EfficientViT executed on the (v) widely-used Edge CPU (Qualcomm Snapdragon 8Gen1) and Edge GPUs, including (vi) NVIDIA Jetson Nano and (vii) NVIDIA Jetson Orin; as well as three SOTA ViT accelerators, including (viii) Auto-ViT-Acc \cite{Li2022AutoViTAccAF} and (ix) Huang et al. \cite{huang2023integer} tailored for standard ViTs and (x) ViA \cite{Wang2022ViAAN} dedicated to Swin Transformer \cite{Liu2021SwinTH} (one of efficient ViTs). We compare them in terms of throughput, energy efficiency, frame rate (FPS), and DSP efficiency.}

\textbf{Accelerator Setup.}
\textbf{\textit{\underline{Characteristics:}}}
The parallelism of computing engines in our accelerator $(N\times M + T\times S)\times L$ (as depicted in Fig. \ref{fig:hw_arch}) is configured to as $(8\times 8 + 8\times 8)\times 16$. Thus, there are a total of $2048$ multipliers in our accelerator, each can execute an $8\times 8$-bit multiplication. To improve DSP utilization, we adopt the SOTA DSP packing strategy \cite{Xilinx-conv} to accommodate two $8$-bit multiplications within each DSP, similar to Auto-ViT-Acc \cite{Li2022AutoViTAccAF} for fair comparisons. 
\textbf{\textit{\underline{Evaluation:}}}
\shh{We implement our accelerator with Verilog, synthesize through Vivado Design Suite, and evaluate on Xilinx ZCU102 FPGA at 200-MHz frequency. {Table \ref{tab:hw_resources} lists our resource consumptions.}}
Furthermore, we follow \cite{Dass2022ViTALiTyUL, p2-vit} to develop a cycle-accurate simulator for our accelerator to obtain fast and reliable estimations and verify them against the RTL implementation to ensure correctness. 

\begin{table}[]
\centering
\caption{Accuracy (\%) comparisons over SOTA PTQ algorithms when weights and activations are both quantized to 8-bit and tested on ImageNet} \vspace{-0.8em}
\setlength{\tabcolsep}{0.2em}
\renewcommand{\arraystretch}{1.2}
\resizebox{\linewidth}{!}{
\begin{threeparttable}{
\begin{tabular}{l|cc|cccc}
\Xhline{3\arrayrulewidth}
\textbf{Method}         & \begin{tabular}[c]{@{}c@{}}\textbf{Softmax}\\ \textbf{Quant}\end{tabular}                                                                        & \textbf{LinAttn}                                                                    & \begin{tabular}[c]{@{}c@{}}\textbf{DeiT-Tiny}\\ -\textbf{R224* \cite{deit}}\end{tabular}       & \begin{tabular}[c]{@{}c@{}}\textbf{DeiT-}\textbf{Small}\\ \textbf{-R224 \cite{deit}}\end{tabular}      & \begin{tabular}[c]{@{}c@{}}\textbf{DeiT-Base}\\ \textbf{-R224 \cite{deit}}\end{tabular}       & \begin{tabular}[c]{@{}c@{}}\textbf{ViT-Base}\\ \textbf{-R224  \cite{vit}}\end{tabular}        \\ \hline \hline
Param. (M)     & --                                                                             & --                                                                         & 5.7                                                             & 22.1                                                            & 86.6                                                            & 86.6                                                            \\
GFLOPs         & --                                                                             & --                                                                         & 1.3                                                             & 4.6                                                             & 17.6                                                            & 17.6                                                            \\ \hline
\rowcolor{dark-blue!16} Full Precision & \color{purple}{\xmark}                       & \color{purple}{\xmark}                   & 72.21                                   & 79.85                                   & 81.85                                   & 84.53                                   \\ \hline
\rowcolor{n2!30} Base PTQ       & \color{purple}{\xmark}                       & \color{purple}{\xmark}                   & 71.78                                                           & 79.35                                                           & 81.37                                                           & 83.48                                                           \\ \hline
Bit-Split \cite{bitsplit}      &                                                                                &                                                                            & -                                                               & 77.06                                                           & 79.42                                                           & -                                                               \\
EasyQuant\cite{Wu2020EasyQuantPQ}      & \multirow{-2}{*}{\color{purple}{\xmark}}     & \multirow{-2}{*}{\color{purple}{\xmark}} & -                                                               & 76.59                                                           & 79.36                                                           & -                                                               \\ \hline
MinMax         &                                                                                &                                                                            & 70.94                                                           & 75.05                                                           & 78.02                                                           & 23.64                                                           \\
EMA \cite{EMA}           &                                                                                &                                                                            & 71.17                                                           & 75.71                                                           & 78.82                                                           & 30.3                                                            \\
Percentile\cite{Percentile}     &                                                                                &                                                                            & 71.47                                                           & 76.57                                                           & 78.37                                                           & 46.69                                                           \\
OMSE \cite{omse}          &                                                                                &                                                                            & 71.3                                                            & 75.03                                                           & 79.57                                                           & 73.39                                                           \\
\rowcolor{dark-green!16} FQ-ViT \cite{Lin2021FQViTPQ}        & \multirow{-5}{*}{\color{dark-green}{\cmark}} & \multirow{-5}{*}{\color{purple}{\xmark}} & 71.07                                                           & 78.4                                                            & 80.85                                                           & 82.68                                                           \\ \hline \hline
\textbf{Method}         & \textbf{Softmax}                                                                        & \textbf{LinAttn}                                                                    & \begin{tabular}[c]{@{}c@{}}\textbf{EfficientViT}\\ \textbf{-B1-R224 \cite{cai2022efficientvit}}\end{tabular} & \begin{tabular}[c]{@{}c@{}}\textbf{EfficientViT}\\ \textbf{-B1-R256 \cite{cai2022efficientvit}}\end{tabular} & \begin{tabular}[c]{@{}c@{}}\textbf{EfficientViT}\\ \textbf{-B1-R288\cite{cai2022efficientvit}}\end{tabular} & \begin{tabular}[c]{@{}c@{}}\textbf{EfficientViT}\\ \textbf{-B2-R224\cite{cai2022efficientvit}}\end{tabular} \\ \hline \hline
Param. (M)     & --                                                                             & --                                                                         & 9.1                                                             & 9.1                                                             & 9.1                                                             & 24                                                              \\
GFLOPs         & --                                                                             & --                                                                         & 0.52                                                            & 0.68                                                            & 0.86                                                            & 1.6                                                             \\ \hline
\rowcolor{dark-blue!16} Full Precision & \color{dark-green}{\xmark}                                                                             & \color{dark-green}{\cmark}               & 79.39                                                           & 79.92                                                           & 80.41                                                           & 82.10                                                           \\ \hline
\rowcolor{n1!25} Base PTQ       & \color{dark-green}{\xmark}                                                                             & \color{dark-green}{\cmark}               & NaN                                                             & NaN                                                             & NaN                                                             & NaN                                                             \\ \hline
\rowcolor{dark-green!16} \textbf{Ours}           & {\color{dark-green}{\xmark}}                                                           & \color{dark-green}{\cmark}               & \textbf{78.64}                                                           & \textbf{78.93}                                                           & \textbf{79.58}                                                           & \textbf{80.97}                                                           \\ \Xhline{3\arrayrulewidth}
\end{tabular}} 
\begin{tablenotes}
		\footnotesize
		\item[*] R224 denotes the resolution of input images is $224\times224$, and so on.
	  \end{tablenotes} 
\end{threeparttable}}\label{tab:alg_results} \vspace{-1em}
\end{table}
\vspace{-0.8em}
\subsection{Evaluation of Trio-ViT's Post-Training Quantization}
\label{sec:exp_alg}
\textbf{Results and Analysis.}
From Table \ref{tab:alg_results}, we can draw \textit{\textbf{four}} conclusions.
\underline{\textbf{(i)}} 
The SOTA post-training quantization (PTQ) method FQ-ViT \cite{Lin2021FQViTPQ}, which develops dedicated quantization schemes to fully quantize all operations in standard ViTs (including Softmax) to enhance hardware efficiency, suffers from $\downarrow$$\mathbf{1.36}\%$ accuracy compared to the full-precision models.
Besides, it also yields an average $\downarrow$$0.75\%$ accuracy compared to the base PTQ, where Softmax and other non-linear operations are not quantized and thus incur non-negligible hardware costs.
This demonstrates that the hardware-unfriendly non-linear operations are sensitive to quantization, hindering both the achievable hardware efficiency and quantization accuracy of standard ViTs. 
\underline{\textbf{(ii)}} To address this limitation, the SOTA efficient ViT dubbed EfficientViT \cite{cai2022efficientvit} has been proposed, which features Softmax-free linear attention (LinAttn) and can achieve much higher accuracy with even fewer parameters and computational costs. This underscores EfficientViT's superiority, highlighting the need for quantization to facilitate its real-world applications.
\underline{\textbf{(iii)}} However, due to the distinct distributions of activation in MBConvs and MSAs, as introduced in Sec. \ref{sec: observations}, the vanilla PTQ method fails to quantize EfficientViT and even yields a Not-a-Number (NaN) issue.
\underline{\textbf{(iv)}} To solve this issue, we propose our dedicated PTQ engine, which can effectively quantize EfficientViTs with merely an average $\downarrow$$\mathbf{0.92}\%$ accuracy when compared with the full precision counterparts, demonstrating our effectiveness.

\begin{table}[]
\centering
\caption{Ablation studies of our post-training quantization engine in terms of proposed channel-wise (CW) migration, filter-wise (FW) shifting, and log2 quantization on ImageNet classification} \vspace{-0.8em}
\setlength{\tabcolsep}{0.4em}
\renewcommand{\arraystretch}{1.2}
\resizebox{\linewidth}{!}{
\begin{threeparttable}{
\begin{tabular}{ccc|cc|cc}
\Xhline{3\arrayrulewidth}
\multicolumn{3}{c|}{\textbf{MBConv Quant}}                                  & \multicolumn{2}{c|}{\textbf{MSA Quant}} & \multirow{2}{*}{\textbf{\begin{tabular}[c]{@{}c@{}}EfficientViT\\ -B1-R224\end{tabular}}} & \multirow{2}{*}{\textbf{\begin{tabular}[c]{@{}c@{}}EfficientViT\\ -B2-R224\end{tabular}}} \\ \cline{1-5}
\textbf{Vanilla}              & \textbf{CW Migration} & \textbf{FW Shifting} & \textbf{Uniform (8)}    & \textbf{Log2 (4)*}    &                                                                                           &                                                                                           \\ \hline \hline
\rowcolor{dark-blue!16} --                            & --                  & --                   & --                  & --               & 79.39                                                                                     & 82.10                                                                                     \\ \hline
\rowcolor{n1!25} \color{dark-green}{\cmark}                           &                   &                    & \color{dark-green}{\cmark}                  &                & NaN                                                                                       & NaN                                                                                       \\ \hline
                           \color{dark-green}{\cmark} &                  &                    & --                  &  --              & \multicolumn{1}{c}{3.23}                                                                      & \multicolumn{1}{c}{0.68}                                                                      \\
                            & \color{dark-green}{\cmark}                 &                    & --                  &  --              & \multicolumn{1}{c}{7.51}                                                                      & \multicolumn{1}{c}{78.52}                                                                      \\
                            &                   & \color{dark-green}{\cmark}                  & --                  &   --             & \multicolumn{1}{c}{28.75}                                                                      & \multicolumn{1}{c}{0.94}                                                                      \\
                            & \color{dark-green}{\cmark}                 & \color{dark-green}{\cmark}                  &  --                 &  --              & \multicolumn{1}{c}{79.05}                                                                      & \multicolumn{1}{c}{81.36}                                                                      \\ \hline
\multicolumn{1}{c}{}          & \color{dark-green}{\cmark}                 & \color{dark-green}{\cmark}                  & \color{dark-green}{\cmark}                 &                & NaN                                                                                       & NaN                                                                                       \\
\rowcolor{dark-green!16} \multicolumn{1}{c}{\textbf{}} & \textbf{\color{dark-green}{\cmark}}        & \textbf{\color{dark-green}{\cmark}}         & \textbf{}           & \textbf{\color{dark-green}{\cmark}}     & \textbf{78.64}                                                                            & \textbf{80.97}  \\ \Xhline{3\arrayrulewidth}                                                                         
\end{tabular}} 
\begin{tablenotes}
		\footnotesize
		\item[*] denotes the 8-bit uniform quantization and 4-bit log2 quantization.
	  \end{tablenotes} 
\end{threeparttable}}\label{tab:alg-ablation} \vspace{-1em}
\end{table}
\textbf{Effectiveness of Our Dedicated Quantization Engine.}
As shown in Table \ref{tab:alg-ablation}, we can see that:
As for the quantization \underline{{\textit{within MBConvs}}}, \underline{\textbf{(i)}} due to the inter-channel variations in DW's inputs and inter-channel asymmetries in PW2's inputs, as introduced in Sec. \ref{sec:observe-mbconv}, vanilla uniform quantization fails to quantize MBConvs in EfficientViT. \underline{\textbf{(ii)}} By incorporating our channel-wise migration and filter-wise shifting, which are proposed to solve the above two issues, respectively, we can effectively quantize MBConvs with only {an average $\downarrow$$0.54\%$} accuracy.
On top of this, regarding the quantization of \underline{{\textit{lightweight MSA}}}, \underline{\textbf{(iii)}} owing to the extreme quantization sensitivity of smaller values in divisors, as illustrated in Sec. \ref{sec:log-observe}, the vanilla $8$-bit uniform quantization yields a NaN issue. 
\underline{\textbf{(iv)}} Thus, we advocate adopting log2 quantization for divisors, which assigns more bins to smaller values and is inherently compatible with the algorithmic property of divisors, thus allowing for quantizing divisors with a mere $4$-bit.  

\begin{table}[]
\centering
\setlength{\tabcolsep}{6pt}
\caption{Accuracy comparisons between vanilla channel-wise (cw) quantization and our proposed channel-wise (cw) migration} \vspace{-0.8em}
\renewcommand{\arraystretch}{1.2}
\resizebox{\linewidth}{!}{
\begin{tabular}{c|cccc}
\Xhline{3\arrayrulewidth}
\textbf{\begin{tabular}[c]{@{}c@{}}Quantization for \\ DW's Inputs\end{tabular}} & \textbf{\begin{tabular}[c]{@{}c@{}}EfficientViT\\ -B1-R224\end{tabular}} & \textbf{\begin{tabular}[c]{@{}c@{}}EfficientViT\\ -B1-R256\end{tabular}} & \textbf{\begin{tabular}[c]{@{}c@{}}EfficientViT\\ -B1-R288\end{tabular}} & \textbf{\begin{tabular}[c]{@{}c@{}}EfficientViT\\ -B2-R224\end{tabular}} \\ \hline \hline
CW Quantization                                                                  & 69.25                                                                    & 78.83                                                                    & 79.36                                                                    & 79.98                                                                    \\
\rowcolor{dark-green!16} \textbf{CW Migration}                                                              & \textbf{78.64}                                                           & \textbf{78.93}                                                           & \textbf{79.58}                                                           & \textbf{80.97}                                                           \\ 
\textbf{\textcolor{dark-green}{Improve (\%)}}                                                            & \textbf{\textcolor{dark-green}{$\uparrow$9.39}}                                                            & \textbf{\textcolor{dark-green}{$\uparrow$0.10}}                                                            & \textbf{\textcolor{dark-green}{$\uparrow$0.22}}                                                            & \textbf{\textcolor{dark-green}{$\uparrow$0.99}}        \\ \Xhline{3\arrayrulewidth}                                                   
\end{tabular}}  \label{tab:alg_cw_scaling} \vspace{-0.5em}
\end{table}
\textbf{Effectiveness of Proposed Channel-Wise Migration.}
Considering the unique algorithmic property of DWConvs, where each channel of weights serves as an independent filter to process each input channel, channel-wise quantization serves as a straightforward solution to solve the inter-channel variations in DW's input, as explained in Sec. \ref{sec: cw-qaunt}.
However, it will greatly increase the number of scaling factors, challenging quantization optimization via LSQ \cite{Esser2019LearnedSS} and limiting achievable accuracy.
Thus, we advocate adopting channel-wise migration on top of layer-wise quantization.
As validated in Table \ref{tab:alg_cw_scaling}, by doing this, we offer an average $\uparrow$$\mathbf{2.67\%}$ accuracy, demonstrating our superiority in solving the inter-channel variations of DW's input while maintaining optimization efficiency.

\begin{table}[]
\caption{{\shh{Generalization of our post-training quantization engine on Semantic Segmentation and tested on the Cityscapes dataset}} \vspace{-0.8em}}
\setlength{\tabcolsep}{0.4em}
\renewcommand{\arraystretch}{1.2}
\resizebox{\linewidth}{!}{
\begin{tabular}{c|ccc|cc|c}
\Xhline{3\arrayrulewidth}
\multirow{2}{*}{\textbf{Methods}}                                                             & \multicolumn{3}{c|}{\textbf{MBConv Quant}}                        & \multicolumn{2}{c|}{\textbf{MSA Quant}} & \multirow{2}{*}{\textbf{\begin{tabular}[c]{@{}c@{}}EfficientViT\\ -B0-R1024\end{tabular}}} \\ \cline{2-6}
                                                                                              & \textbf{Vanilla}     & \textbf{CW Migration} & \textbf{FW Shifting} & \textbf{Uniform (8)}     & \textbf{Log2 (4)}    &                                                                                            \\ \hline \hline
\rowcolor{dark-blue!16} \textbf{Full Precision}                                                                       & --                   & --                  & --                   & --                   & --               & 75.65                                                                                      \\ \hline
\rowcolor{n1!25} \textbf{Base PTQ}                                                                            & \textbf{\color{dark-green}{\cmark}}                  &                   &                    & \textbf{\color{dark-green}{\cmark}}                  &                & N/A                                                                                        \\ \hline
\multirow{4}{*}{\textbf{\begin{tabular}[c]{@{}c@{}}Ablation Study \\ in MBConv\end{tabular}}} & \textbf{\color{dark-green}{\cmark}}                  &                   &                    & --                   & --               & 36.89                                                                                      \\
                                                                                              &                    & \textbf{\color{dark-green}{\cmark}}                 &                    & --                   & --               & 72.51                                                                                      \\
                                                                                              &                    &                   & \textbf{\color{dark-green}{\cmark}}                  & --                   & --               & 29.02                                                                                      \\
                                                                                              &                    & \textbf{\color{dark-green}{\cmark}}                 & \textbf{\color{dark-green}{\cmark}}                  & --                   & --               & 75.22                                                                                      \\ \hline
\multirow{2}{*}{\textbf{\begin{tabular}[c]{@{}c@{}}Ablation Study\\  in MSA\end{tabular}}}    &   & \textbf{\color{dark-green}{\cmark}}                 & \textbf{\color{dark-green}{\cmark}}                  & \textbf{\color{dark-green}{\cmark}}                  &                & 1.12                                                                                       \\
                                                                                              & \cellcolor{dark-green!16}{}   &\cellcolor{dark-green!16} \textbf{\color{dark-green}{\cmark}}                 & \cellcolor{dark-green!16} \textbf{\color{dark-green}{\cmark}}                  & \cellcolor{dark-green!16}                     & \cellcolor{dark-green!16} \textbf{\color{dark-green}{\cmark}}              & \cellcolor{dark-green!16}\textbf{74.83}                                                                             \\ \Xhline{3\arrayrulewidth}
\end{tabular}} \label{table:alg_seg}
\end{table}
\shh{\textbf{Generalization on Semantic Segmentation.}
To assess the generalization capability of our proposed quantization method, we apply it to EfficientViT-B0-R1024 \cite{cai2022efficientvit} and evaluate its performance on the semantic segmentation task, using the Cityscapes \cite{cordts2016cityscapes} as the dataset and mean Intersection over Union (mIoU) as the evaluation metric. As listed in Table \ref{table:alg_seg}, we can see that:
For the quantization \underline{{\textit{within MBConvs}}}, vanilla uniform quantization fails due to the inter-channel variations in DW's inputs and inter-channel asymmetries in PW2's inputs. By combining our channel-wise (CW) migration and filter-wise (FW) shifting, we achieve effective quantization of MBConvs with only a $0.33\%$ reduction in mIoU. Note that, due to the interdependence between layers, while filter-wise shifting alone is inferior to the vanilla approach, it can boost performance when incorporated with our channel-wise migration, proving its effectiveness and necessity.
Furthermore, regarding the quantization of \underline{{\textit{lightweight MSA}}}, owing to the extreme quantization sensitivity of smaller values in divisors, the vanilla $8$-bit uniform quantization yields a catastrophic performance drop. 
In contrast, our proposed log2 quantization enables the effective quantization of divisors with a mere $4$-bit. This set of results demonstrates the generalization ability and robustness of our quantization method.}


\begin{figure}[t]
	\centerline{\includegraphics[width=\linewidth]{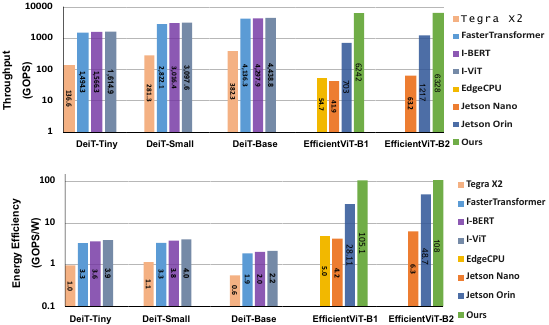}}
	\vspace{-0.8em}
	\caption{Results comparisons with SOTA baselines on GPUs/CPU in terms of throughput and energy efficiency. Input resolution is $224\times 224$ here. {\textbf{Note that}} the y-axis is plotted on a logarithmic scale for better illustration.} 
	\label{fig:hw_results} \vspace{-1.2em}
\end{figure} 
\vspace{-0.6em}
\subsection{Evaluation of Trio-ViT's Dedicated Accelerator}
\label{sec:exp_hw}
\textbf{Comparisons with GPUs/CPU.}
We follow \cite{Ham2021ELSAHC, Qu2022DOTADA} to scale up the hardware resources of our accelerator to have comparable peak throughput with the general computing platform (i.e., NVIDIA 2080Ti GPU) to enable fair comparisons with GPUs/CPU. As shown in Fig. \ref{fig:hw_results} (where the y-axis is plotted on a \textbf{\textit{logarithmic}} scale for better illustration), we can achieve much better hardware efficiency compared to SOTA baselines on GPUs/CPU, validating our effectiveness. Specifically, \underline{\textbf{(i)}} when compared with the full-precision DeiTs \cite{deit} on Edge GPU (Tegra X2), we can achieve $\uparrow$$17\times$$\sim$$\uparrow$$46\times$ and $\uparrow$$92\times$$\sim$$\uparrow$$195\times$ throughput and energy efficiency, respectively.
\underline{\textbf{(ii)}} For comparisons with 8-bit DeiTs quantized by FasterTransformer \cite{fastertransformer}, I-BERT \cite{Kim2021IBERTIB} and I-ViT \cite{Li2022IViTIQ}, and executed on the Turning Tensor Core of NVIDIA 2080Ti GPU, we can offer $\uparrow$$1.4\times$$\sim$$\uparrow$$4.2\times$ and $\uparrow$$26\times$$\sim$$\uparrow$$58\times$ throughput and energy efficiency, respectively.
Furthermore, \underline{\textbf{(iii)}} regarding the comparison with full precision EfficientViT \cite{cai2022efficientvit} on edge CPU, we can gain up to $\uparrow$$116\times$ and $\uparrow$$22\times$ throughput and energy efficiency, respectively.
\underline{\textbf{(iv)}} When compared with EfficientViT on edge GPUs (NVIDIA Jetson Nano and	Jetson Orin), we can gain $\uparrow$$5.2\times$$\sim$$\uparrow$$149\times$ and $\uparrow$$2.2\times$$\sim$$\uparrow$$25\times$ in terms of throughput and energy efficiency.

\begin{table}[t]
\centering
\setlength{\tabcolsep}{0.5pt}
\caption{\shh{Comparisons with SOTA ViT accelerators}} \vspace{-0.3em}
\renewcommand{\arraystretch}{1.2}
\resizebox{\linewidth}{!}{
\begin{tabular}{c|c|cc|cc|ccc}
\Xhline{3\arrayrulewidth}
\textbf{Accelerator}                                                            & \textbf{Via} \cite{Wang2022ViAAN}                                 & \multicolumn{2}{c|}{\textbf{Huang et al.} \cite{huang2023integer}}                       & \multicolumn{2}{c|}{\textbf{Auto-ViT-Acc} \cite{Li2022AutoViTAccAF}}                                                                                               & \multicolumn{3}{c}{\textbf{\textcolor{dark-green}{Ours}}}                                                                                                                                                                                              \\ \hline \hline
\textbf{Device}                                                                 & \textbf{\begin{tabular}[c]{@{}c@{}}Xilinx\\ Alveo U50\end{tabular}} & \multicolumn{2}{c|}{\textbf{\begin{tabular}[c]{@{}c@{}}Xilinx\\ ZCU102\end{tabular}}} & \multicolumn{2}{c|}{\textbf{\begin{tabular}[c]{@{}c@{}}Xilinx\\ ZCU102\end{tabular}}}                                                    & \multicolumn{3}{c}{\textbf{\begin{tabular}[c]{@{}c@{}}Xilinx\\ ZCU102\end{tabular}}}                                                                                                                                           \\ \hline
\textbf{Frequency (MHz)}                                                        & \textbf{300}           & \multicolumn{2}{c|}{\textbf{300}}                                               & \multicolumn{2}{c|}{\textbf{150}}                                                                                                        & \multicolumn{3}{c}{\textbf{200}}                                                                                                                                                                                               \\ \hline
\textbf{Format}                                                                 & \textbf{FP16}            & \multicolumn{2}{c|}{\textbf{INT8}}                                           & \multicolumn{2}{c|}{\textbf{INT8}}                                                                                                       & \multicolumn{3}{c}{\textbf{INT8}}                                                                                                                                                                                              \\ \hline
\textbf{Model}                                                                  & \textbf{\begin{tabular}[c]{@{}c@{}}Swin-T\\ -R224\end{tabular}}   & \textbf{\begin{tabular}[c]{@{}c@{}}ViT-T\\ -R224\end{tabular}} & \textbf{\begin{tabular}[c]{@{}c@{}}ViT-S\\ -R224\end{tabular}}   & \textbf{\begin{tabular}[c]{@{}c@{}}DeiT-S\\ -R224\end{tabular}} & \textbf{\begin{tabular}[c]{@{}c@{}}DeiT-B\\ -R224\end{tabular}} & \textbf{\begin{tabular}[c]{@{}c@{}}Effi.ViT\\ -B1-R288\end{tabular}} & \textbf{\begin{tabular}[c]{@{}c@{}}Effi.ViT\\ -B2-R224\end{tabular}} & \textbf{\begin{tabular}[c]{@{}c@{}}Effi.ViT\\ -B2-R256\end{tabular}} \\ \hline
\textbf{DSP Used}                                                               & 2420       & {{1268}}  & {{1268}}                                                          & 1936                                                                & 2066                                                               & 1024                                                                     & 1024                                                                     & 1024                                                                     \\
\textbf{Latency (ms)}                                                           & 14.5   & 4.1  & 11.2                                                              & 12.8                                                                & 38.6                                                               &   \textbf{\textcolor{dark-green}{2.24}}                                                            & \textbf{\textcolor{dark-green}{4.05}}                                                            & \textbf{\textcolor{dark-green}{5.28}}                                                            \\
\textbf{Frame Rate (FPS)}                                                       & 68.8           & 245	& 89.3                                                     & 78.1                                                                & 25.9                                                               &   \textbf{\textcolor{dark-green}{447}}                                                             & \textbf{\textcolor{dark-green}{247}}                                                             & \textbf{\textcolor{dark-green}{190}}                                                             \\ \hline
\textbf{\begin{tabular}[c]{@{}c@{}}Throughput\\ \textbf{(GOPS)}\end{tabular}}            & 310       & 616	& 763                                                          & 711                                                                 & 900                                                                &  \textbf{\textcolor{dark-blue}{769}}                                                                      & \textbf{\textcolor{dark-blue}{791}}                                                                      & \textbf{\textcolor{dark-blue}{796}}                                                                      \\ \hline
\textbf{\begin{tabular}[c]{@{}c@{}}DSP Efficiency\\(GOPS/DSP)\end{tabular}} & 0.13     & 0.49	& 0.60                                                           & 0.37                                                                & 0.44                                                               &   \textbf{\textcolor{dark-green}{0.75}}                                                            & \textbf{\textcolor{dark-green}{0.77}}                                                            & \textbf{\textcolor{dark-green}{0.78}}                                                            \\ \hline
\textbf{\begin{tabular}[c]{@{}c@{}}Energy Efficiency\\ (GOPS/W)\end{tabular}}   & 7.92    & --	& 25.8                                                            & 84.1                                                                & 95.7                                                               &   \textbf{\textcolor{dark-green}{105}}                                                             & \textbf{\textcolor{dark-green}{108}}                                                             & \textbf{\textcolor{dark-green}{109}}                                                             \\ \hline
\textbf{Accuracy (\%)}                                                          & 81.3         & 74.39	& --                                                       & 79.69                                                               & 81.93                                                              & \textbf{79.58}                                                                    & \textbf{80.97}                                                                    & \textbf{81.62}      \\ \Xhline{3\arrayrulewidth}                                               
\end{tabular}} \label{tab:hw-accelerator-results} \vspace{-1.5em}
\end{table}

\textbf{Comparisons with SOTA ViT Accelerators.}
From Table \ref{tab:hw-accelerator-results}, we can see that:
\underline{\textbf{(i)}} Due to the superiority of our quantization engine and the promising hardware efficiency of EfficientViT compared to standard ViTs/DeiTs, we gain the lowest latency and highest FPS under comparable accuracy. \shh{Particularly, we achieve up to $\uparrow$$\textbf{3.6}\times$ FPS compared to Via \cite{Wang2022ViAAN}, a dedicated accelerator for an efficient ViT dubbed Swin-Transformer-Tiny (Swin-T) \cite{Liu2021SwinTH}, as well as $\uparrow$$\mathbf{5.0\times}$ and $\uparrow$$\mathbf{7.3\times}$ FPS over dedicated ViT/DeiT accelerators Huang et al. \cite{huang2023integer} and Auto-ViT-Acc \cite{Li2022AutoViTAccAF}, respectively.
\underline{\textbf{(ii)}} Furthermore, owing to the hybrid design of our dedicated accelerator, which is proposed to effectively support various operators in the Convolution-Transformer hybrid architecture of EfficientViT with enhanced hardware utilization, we can achieve the highest hardware utilization efficiency. For example, we can offer up to $\uparrow$$\mathbf{6.0}\times$, $\uparrow$$\mathbf{1.5}\times$, $\uparrow$$\mathbf{2.1}\times$ DSP efficiency when compared with Via, Huang et al., and Auto-ViT-Acc, respectively.
\underline{\textbf{(iii)}} Additionally, thanks to our developed pipeline architecture aiming to facilitate both inter- and intra-layer fusion, we gain the best {energy efficiency, i.e., up to $\uparrow$$\mathbf{13.7}\times$, $\uparrow$$\mathbf{4.1}\times$, and $\uparrow$$1.3\times$} compared to them.}

\section{Conclusion}
\label{sec:conclusion}
In this paper, we have proposed, developed, and validated Trio-ViT, the first post-training quantization and acceleration framework dedicated to the state-of-the-art (SOTA) efficient Vision Transformer (ViT), dubbed EfficientViT.
Specifically, at the algorithm level, we propose a tailored post-training quantization engine that incorporates several innovative quantization schemes to effectively quantize EfficientViT with enhanced quantization accuracy.
At the hardware level, we develop a dedicated accelerator integrating a hybrid design and a pipeline architecture to boost hardware efficiency.
Extensive experimental results consistently prove our effectiveness. Particularly, we gain up to $\uparrow$$\mathbf{3.6}\times$, $\uparrow$$\mathbf{5.0}\times$, and $\uparrow$$\mathbf{7.3}\times$ FPS with comparable accuracy over SOTA ViT accelerators.

\textbf{Limitations and Future Work.} 
It has been widely demonstrated that model layers exhibit varying degrees of sensitivity to quantization, thus allocating the same bit to all layers is deemed sub-optimal in both accuracy and efficiency \cite{Liu2021PostTrainingQF, Dong2019HAWQHA}. Therefore, our future research will focus on exploring mixed quantization, considering variations in both quantization bits and schemes (such as fix-point and power-of-two).

\bibliographystyle{unsrt}
\bibliography{main}

\end{document}